\documentclass[11pt]{article}

\usepackage[preprint]{acl}

\usepackage{times}
\usepackage{latexsym}
\usepackage{subcaption}
\usepackage{multirow}
\usepackage{amssymb}
\usepackage{fontawesome5}
\usepackage{tcolorbox} 
\usepackage{xcolor} 
\usepackage{fvextra} 
\DefineVerbatimEnvironment{Verbatim}{Verbatim}{breaklines=true}

\usepackage[T1]{fontenc}

\usepackage[utf8]{inputenc}

\usepackage{microtype}

\usepackage{inconsolata}

\usepackage{graphicx}
\usepackage{booktabs}
\usepackage{amsmath}
\usepackage{longtable}

\title{Computational Representations of Character Significance in Novels}

\author{
  Haaris Mian\thanks{Equal contribution} \\
  Applied Mathematics \\
  Columbia University \\
  \texttt{ham2176@columbia.edu}
  \And
  Melanie Subbiah\footnotemark[1] \\
  Computer Science \\
  Columbia University \\
  \texttt{mss2290@columbia.edu}
  \And
  Sharon Marcus \\
  English \& Comparative Literature \\
  Columbia University \\
  \texttt{sm2247@columbia.edu}
  \AND
  Nora Shaalan \\
  English \& Comparative Literature \\
  Columbia University \\
  \texttt{ns3486@columbia.edu}
  \And
  Kathleen McKeown \\
  Computer Science \\
  Columbia University \\
  \texttt{kathy@cs.columbia.edu}
}

\begin{document}
\maketitle
\begin{abstract}
Characters in novels have typically been modeled based on their presence in scenes in narrative—considering aspects like their actions, named mentions, and dialogue. This conception of character places significant emphasis on the main character who is present in the most scenes. In this work, we instead adopt a framing developed from a new literary theory proposing a six-component structural model of character. This model enables a comprehensive approach to character that accounts for the narrator / character distinction and includes a component neglected by prior methods, discussion by other characters.
We compare general-purpose LLMs with task-specific transformers
for operationalizing this model of character on major 19th-century British realist novels. 
Our methods yield both component-level and graph representations of character discussion. 
We then demonstrate that these representations allow us to approach literary questions at scale from a new computational lens. Specifically, we explore Woloch's classic ``the one vs the many" theory of character centrality and the gendered dynamics of character discussion.

\end{abstract}

\section{Introduction}

When reading a novel, we typically come away with a sense of who the main character is, but even novels with singular lead protagonists have many other significant characters. Traditional literary theory classifies characters as major and minor \cite{Forster1927, Woloch2003}, but many characters blur this distinction. 
For a given character, we aim to quantify how much ``space" the character occupies in the story, thereby quantifying their significance. 

A substantial portion of computational work on character has focused on social networks extracted from the text through character co-occurrence \cite{amalvy-etal-2025-role} or dialogue exchange \cite{elson-etal-2010-extracting}. While these analyses are useful, they construct character significance almost exclusively through interaction: a character takes up space through co-occurrence or dialogue with others. This idea is somewhat contradictory to  Woloch's prominent literary theory of "the one vs. the many" \cite{Woloch2003}. Woloch argues that narratives are ``structured around the relationship between one central individual who dominates the story and a host of subordinate figures who jostle for, and within, the limited space that remains." This frames character prominence as a zero-sum game; other characters must be suppressed to elevate the central figure \cite{schaffer2021communities}. 

We collaborate with researchers in English and Comparative Literature who have developed a six-component model of individual character significance in novels. These components bring into focus how characters take up space beyond physical presence in scenes. For example, \textit{interiority} is a component measuring the expression of the character's voiced and unvoiced thoughts and feelings. There is also a component measuring mentions of a character in dialogue from other characters (\textit{discussion by other characters}), which foregrounds other characters' interest in them. We use this component-based definition to develop computational representations of character significance. 


In addition, we leverage the \textit{discussion by other characters} component to develop a new type of social graph. 
We explore how discussion by other characters forms a directed ``discussion" graph, which can bring a new view to character centrality in novels. 
We use this representation to answer literary analytic questions about the gendered aspects of character discussion. We also compare 
count-based measures of character significance with network-based measures and highlight how varying definitions of taking up narrative ``space" lead to differing rankings of prominence in the narrative. 

The core contributions of our work are:
\begin{itemize}
\vspace{-0.2cm}
\item Framing character representations with a six-component model of character, developed in literary theory \cite{Marcus2025CaringCharacter}
and releasing an accompanying dataset of expert-annotated component scores.
\vspace{-0.2cm}
\item Developing and releasing a pipeline that 
can accurately label these components across 19th-century novels\footnote{Data and code available at \url{github.com/haarisamian/tagging-gutenberg}.} 
and create character representations, including component-based frequency counts and a new network representation.
\vspace{-0.2cm}
\item Using these representations to perform a corpus-level analysis addressing questions in literary theory around how characters are centered in novels.
\end{itemize}



\section{Related Work}

\citet{piper-etal-2021-narrative} 
comprehensively categorize the use of NL techniques for literary theory;
our work fits most closely within the category of {\em agents}, 
foregrounding the centrality of characters.
Recently, 
\citet{thai-iyyer-2025-literary} investigate how well LLMs understand literary texts by prompting them to retrieve quotations from a novel.
They find that while large, closed source models do well overall, all models struggle with difficult scenarios. 
Their work clearly shows that further modeling work is required if LLMs are to perform well on literary analysis tasks. Our work, like \citet{thai-iyyer-2025-literary}, examines whether LLMs can be used to successfully carry out this analysis. 

Unsurprisingly, given the importance of characters in literary theory, 
other research focuses solely on detection of characters and their attributes \cite{bamman-etal-2014-bayesian,  rashkin-etal-2018-modeling, piper-etal-2024-social}.
Work by \citet{amalvy-etal-2025-role} assesses cascading, pipelined methods for identifying characters, 
demonstrating the need for better coreference approaches.
They find that traditional pipelined methods outperform LLMs in terms of recall. 
Other work 
\cite{soni-etal-2023-grounding} focuses on grounding characters to their narrative locations and uses these results to analyze literary hypotheses (e.g., showing that protagonists are more mobile than minor characters).
This line of work has yielded a valuable resource for identifying characters in the form of a corpus of character coreference annotations~\cite{bamman-etal-2020-annotated}. 

Additionally, there is a substantial body of work extracting and analyzing character networks from literary texts. \citet{labatut2019extraction} provide a survey of character network extraction and analysis methods. Foundational work by \citet{moretti2011network} introduced network analysis as a tool for literary criticism, using manually annotated networks to analyze narrative structure in Hamlet.
Subsequent work has automated network construction through co-occurrence \citep{bonato2016mining} and dialogue detection \citep{evalyn2018analyzing, elson-etal-2010-extracting},
enabling analysis at larger scales \citep{hamilton-etal-2025-city}. A key finding is that character networks often exhibit small-world and scale-free properties \citep{li2019complex}, a phenomenon common to both literary and other natural social networks.

Centrality measures have been used to identify protagonists and analyze character prominence \citep{beveridge2016network, masias2017exploring, agarwal-etal-2012-social}. Our work extends this research by grounding network analysis in a component-based theory of character, more specifically the relational ``discussion" component we propose in Section 3, which offers a new line of analysis not covered by prior co-occurrence or dialogue approaches. While \citet{sims-bamman-2020-measuring} investigate information propogation in social networks and their gendered dynamics, no principled investigation has been done of the propagation of information specifically focusing on the discussion of other characters and the gendered dynamics of such discussion as we present here.

\section{Narrative Components of Character}

\begin{table*}[t]
    \centering \small
    \begin{tabular}{r|cccccc|cccccc}
    \toprule
         & \multicolumn{6}{c|}{\textit{Pride and Prejudice}} & \multicolumn{6}{c}{\textit{Jane Eyre}}\\
         & \textbf{N} & \textbf{A} & \textbf{C} & \textbf{I} & \textbf{DC} & \textbf{DN} & \textbf{N} & \textbf{A} & \textbf{C} & \textbf{I} & \textbf{DC} & \textbf{DN}\\\midrule
         Avg. per chapter& 52.3 & 29.9 & 23.1 & 32.9 & 60.3 & 7.7 & 67.1 & 143.5 & 79.9 & 110.2 & 88.7 & 22.1\\
         Avg. per character& 58.0 & 33.1 & 25.6 & 36.5 & 66.9 & 8.5 & 15.2 & 32.5 & 18.1 & 24.9 & 20.1 & 5.0\\
         Total& 3192 & 1823 & 1409 & 2006 & 3679 & 467 & 2548 & 5453 & 3038 & 4186 & 3371 & 839\\\bottomrule
    \end{tabular}
    \caption{Component score summary statistics across the two hand-counted novels in our corpus.}
    \label{tab:compcounts}
\end{table*}

Our team includes literary researchers, who have developed an original model of character significance. In the past, scholars of character have focused on: 1) Character types (e.g. hero, helper, antagonist) or social roles (e.g., artist, teacher, banker) \cite{Propp1968}, 2) Character prominence (e.g., major vs. minor \cite{Woloch2003}, flat vs. round \cite{Forster1927, Figlerowicz2016}, and 3) Realism \cite{barthes1968reality}. 


The literary researchers on our team have instead developed a structural model of character consisting of six components \cite{Marcus2025CaringCharacter}. While prior studies have focused on the importance of individual components to the construction of character, such as name \cite{margolin2002naming,barthes1974s}, action \cite{piper2024characters}, interiority \cite{eder2010characters}, and direct speech \cite{menon2019keeping, menon2024constructing, menon2024forms}, none have presented an integrated view of these components as is presented here. 
Using examples from \textit{Pride and Prejudice}, we define these components below for a given character:
\par{\textbf{Name (N)}:} instances of the character's proper names. For example, \textit{Elizabeth}, \textit{Lizzy}, and \textit{Miss Bennet} (when used to refer to her) all count as mentions of Elizabeth, but we do not include pronouns or relational mentions.
\par{\textbf{Communication (C)}}: sentences of the character's quoted speech or writing. 
\par{\textbf{Interiority (I)}}: verb phrases expressing what the character thinks, feels, intends, or perceives. For example, ``\textit{Elizabeth found herself quite equal to the scene...}"
\par{\textbf{Action (A)}}: verb phrases conveying the character's actions that are not quoted speech or interiority. For example, ``\textit{Charlotte appeared at the door...}"
\par{\textbf{Discussion by other characters (DC)}}: 
sentences of communication (C) from another character in which this character is discussed. For example, when Elizabeth says of Mr. Darcy, ``\textit{I could easily forgive his pride if he had not mortified mine,}" we consider this a DC of Mr. Darcy (pronouns and relational mentions are included).
\par{\textbf{Description by the narrator (DN)}}: sentences in the narrative voice that describe the character—typically their looks, manner, or dress. For example, ``\textit{Miss de Bourgh was pale and sickly...}"

These six components 
can be used to analyze characters both individually and relative to one another. The larger the sum of a character’s components, the more prominent the character becomes. Two of these components can be relational (C and DC), allowing us to additionally represent them through networks. While network analysis using dialogue networks (component C) has been done before, discussion networks (component DC) are a new representation we propose. 

\section{Dataset}

To work with these components of character, our English and Comparative Literature researchers manually annotate all chapters in \textit{Pride and Prejudice} by Jane Austen and \textit{Jane Eyre} by Charlotte Bronte. \textit{Pride and Prejudice} consists of 61 chapters and 55 characters. \textit{Jane Eyre} consists of 38 chapters and 168 characters. They do not record each span, but keep track of the counts for each component for each character at the chapter level. We use the chapter-level counts from this hand annotation as the source of ground truth to compare automatic methods against.

While this annotation process is too labor-intensive to perform for a large quantity of books, we still obtain thousands of ground truth counts (as shown in Table \ref{tab:compcounts}) given the lengths of the novels. Additionally, in labeling components, \textit{Jane Eyre} introduces different challenges from \textit{Pride and Prejudice} since it has a first-person narrator. Jane is both the central character and the narrator, so components like \textit{interiority} (I) and \textit{description by the narrator} (DN) are more difficult to computationally distinguish.

To present a scaled analysis (on novels for which we do not have ground truth), we create a corpus of 
64 major titles by English, Scottish, and Irish authors, published between 1800 and 1900. We exclude novels with first-person narrators who are also characters, as our analysis of \textit{Jane Eyre} (shown in later sections) demonstrates our methods do not work well for these novels. The titles on this list have been widely read and commented on since their publication, and span a range of sub-genres (see Appendix Table \ref{tab:corpus} for the full list). We obtain these texts from Project Gutenberg using a standard pipeline \cite{gerlach2020standardized}.



\section{Methods}

We consider a novel as a document $D$, containing a set of chapters $C=\{c_1,...,c_n\}$, and associated with a set of characters $K =\{k_1,..., k_n\}$.

\subsection{Labeling components of character}
\label{subsec:component_score}
Our goal is to produce the correct count of tags $V = [t_N, t_C, t_I, t_A, t_{DN}, t_{DC}]$ for each character $k$ in each chapter $C$. $V$ is a representation of the character, with each of these counts serving as a \textit{component score} for the character, and the total of all counts serving as the \textit{total score} for the character. We compare two approaches for automatically tagging each chapter: (i) chapter-level counts, and (ii) span-level annotations.


\par{\textbf{Chapter-level counts.}} In this setting, we give a model a chapter $c$ as input and predict component scores $[t_{N_k}, t_{C_k}, t_{I_k}, t_{A_k}, t_{{DN}_k}, t_{{DC}_k}]$ for each character $k$ as the output. This setting most closely matches our ground truth data which provides chapter-level counts, and many LLMs are capable of generating structured output like this when requested. While highly efficient, this method does not provide any interpretable sense of which spans of text induce the counts.

\par{\textbf{Span-level labeling.}} For counts grounded in text spans, we use parsing-based approaches to process spans of text individually and accumulate tag counts from these spans. For example, for \textit{discussion by other characters} (DC), we first parse out dialogue and then use models to identify whether a dialogue turn mentions another character. 
Having span-level annotations for components like DC allows us to construct network representations.

\subsection{Social network graphs}
Here, we consider a graph-based representation of character relationships. Formally, we aim to recover a graph $G = (V, E)$, where each node $v \in V$ represents a character $k$ in the novel, and each edge $e \in E$ represents interaction between two characters and $W: E\rightarrow\mathbb{R}$ is a suitable weighting on the edges encoding interaction strength. We construct three character networks per novel: 

\par{\textbf{Co-occurrence network}}: Undirected edges count paragraph-level co-presence of characters. 
\par{\textbf{Dialogue network}}: Undirected edges count spoken turn-taking exchanges. These edges correspond to the \textit{communication} (C) component.
\par{\textbf{Discussion network}}: Directed edges $u \rightarrow v$ are weighted by the number of sentences in which character $u$ discusses character $v$ (including direct address and third-party discussion). These edges correspond to the \textit{discussion by other characters} (DC) component.


While co-occurence and dialogue graphs have been studied before, the discussion graph introduces a new perspective on social dynamics. Such a network, we hypothesize, might capture a more nuanced notion of centrality within the novel.



\section{Experimental Setup}

\subsection{Character representations}
For all steps requiring an LLM, we use \texttt{gpt-4o-mini} either in regular prompting mode with temperature 0, or structured output mode. All prompts discussed in this section are shown in Appendix \ref{sec:prompts}.

\par{\textbf{LLM chapter-level counts.}} We use an LLM for predicting chapter-level counts and compare all combinations of the following settings specifying the granularity of the task: full chapter input vs. 1000-token chunks, counting for all characters vs. each character individually, and counting for all components vs. each component individually. In all cases, we prompt the model to directly generate a JSON of counts for each tag for each character in structured output mode.

\par{\textbf{Component-specific span-level labels.}} For span-level labeling, we compare two approaches: (i) a BookNLP-based pipeline, and (ii) LLM span-level annotations. 
The BookNLP package\footnote{\url{https://github.com/booknlp/booknlp}} is a widely used transformer-based pipeline designed for literary texts \cite{bamman-etal-2019-annotated, bamman-etal-2020-annotated,sims-bamman-2020-measuring}. 
We use the \texttt{entity,quote,supersense,event,coref} packages to preprocess the novels. We then parse these output files into the spans that meet the requirements for each component tag. 
In addition to the supersense labeling for verbs, we consult VerbNet classes \citep{schuler2005verbnet} to identify instances of \textit{action} (A) and \textit{interiority} (I). 

In order to standardize the BookNLP character clusters, 
we ask the LLM to map BookNLP character names to our ground truth list of character names for the two books we have ground truth lists for. For novels in our larger corpus, we leverage the BookNLP outputs to construct such a list by filtering the tagged entities by ``person'' mentions and retaining the coreference clusters whose total and proper mentions exceed a desired threshold (chosen as 1 for simplicity). The difficulty of assigning character names to coreference clusters without a proper mention in first-person novels is illustrated in Appendix Figure \ref{fig:jane-ablation}.

For LLM span-level labels, we process each component individually and design prompts specific for that task, which produce component scores and their associated text spans. For example, for the DC component, we first parse out dialogue and then ask the LLM to resolve coreferences to characters within that dialogue turn. We find this method more reliable than asking the LLM directly who is discussed in the dialogue.


\par{\textbf{Metrics.}} For the accuracy of tagging components, we assess the Mean Absolute Error (MAE) and Pearson correlation of each component score (across all characters and chapters) relative to the gold counts. We define MAE as 
\begin{equation*}
\text{MAE} =  \frac{\sum_{c\in C}\frac{\sum_{k\in K}\text{abs}(g_{ck} - p_{ck})}{|K|}}{|C|}
\end{equation*}
where $g_{ck}$ is the gold count for character $k$ in chapter $c$ and $p_{ck}$ is the predicted count for character $k$ in chapter $c$.
The MAE indicates how off the count for each character is, while the correlation indicates whether the relative counts are meaningful even if the exact numbers are not precise. 

\par{\textbf{Character networks.}}
We extract co-occurrence and dialogue networks by first running the novel through BookNLP. For component-based interactions, we utilize the automated tagging methods described earlier to identify instances where one character discusses another (DC component).

 We also assign a binary gender label (M/F) to characters using the majority third-person pronoun in their BookNLP coreference cluster for downstream analysis; this proxy is imperfect and does not capture all the nuances of gender, but reflects the appearance of gendered textual references. 

\subsection{Corpus analysis}
We use various metrics to compare character significance using different character representations across our corpus. For component scores, we can easily identify character significance based on their total or per-component score. For network graphs, we compute centrality measures, including degree, betweenness, closeness, eigenvector centrality, and PageRank (definitions shown in Appendix Table \ref{tab:metrics}). For the directed discussion (DC) graphs, we distinguish between in-degree and out-degree, where the former captures how much a character is talked about by others, and the latter captures how much they talk about others. We additionally quantify the inequality of character significance through global network measures such as the Gini coefficient, degree centralization, and assortativity. Finally, extending the geometric intuitions of prior work on the scale-free structure of character networks, we train and visualize Poincar\'{e} embeddings of graphs across the corpus as described in Appendix \ref{sec:poin}.

\begin{table*}[t]
\centering \small
    \begin{tabular}{ccccccccc|c|cccccc|c}
    \toprule
    \multicolumn{3}{c}{\textbf{Method}} & \multicolumn{7}{c|}{Mean Absolute Error ($\downarrow$)} & \multicolumn{7}{c}{Correlation ($\uparrow$)}\\\midrule
         \tiny \faBook & \tiny \faUser & \tiny \faTag & \textbf{N} & \textbf{A} & \textbf{C} & \textbf{I} & \textbf{DC} & \textbf{DN} & \textbf{avg.} & \textbf{N} & \textbf{A} & \textbf{C} & \textbf{I} & \textbf{DC} & \textbf{DN} & \textbf{avg.}\\\midrule\midrule
&& & 2.28 & 1.98 & 1.48 & 2.11 & 4.14 & 1.08 & 2.18 & 0.75 & 0.66 & 0.67 & 0.74 & 0.47 & 0.49 & 0.57 \\
&&\tiny\checkmark & 2.08 & 2.33 & 1.79 & 2.00 & 4.06 & 1.59 & 2.31 & 0.78 & 0.66 & 0.66 & 0.77 & 0.49 & 0.46 & 0.58 \\
& \tiny\checkmark & & 3.36 & 1.64 & 1.82 & 2.24 & 3.72 & 2.66 & 2.57 & 0.78 & 0.71 & 0.69 & 0.72 & 0.54 & 0.50 & 0.57 \\
&\tiny\checkmark&\tiny\checkmark & 4.0 & 4.75 & 3.36 & 12.48 & 3.92 & 2.67 & 5.20 & 0.78 & 0.65 & 0.66 & 0.8 & 0.53 & 0.46 & 0.53 \\
\tiny\checkmark&& & 3.56 & 1.82 & 1.31 & 1.98 & 3.81 & 1.85 & 2.39 & 0.84 & 0.70 & 0.71 & 0.73 & 0.58 & 0.50 & 0.58 \\
\tiny\checkmark&&\tiny\checkmark & 2.95 & 4.25 & 2.47 & 3.18 & 3.60 & 2.92 & 3.23 & 0.86 & 0.73 & 0.67 & 0.77 & 0.60 & 0.48 & 0.64 \\
\tiny\checkmark&\tiny\checkmark& & 9.09 & 1.60 & 2.38 & 4.16 & 3.54 & 5.14 & 4.32 & 0.80 & 0.73 & 0.67 & 0.71 & 0.58 & 0.48 & 0.57 \\
\tiny\checkmark&\tiny\checkmark&\tiny\checkmark & 4.58 & 12.69 & 4.62 & 26.12 & 5.08 & 4.57 & 9.61 & 0.83 & 0.67 & 0.62 & 0.72 & 0.61 & 0.50 & 0.48 \\
\midrule
\multicolumn{3}{l}{BookNLP} & \textbf{0.83} & 1.92 & 1.16 & 1.95 & 3.68 & 0.67 & 1.70 & \textbf{0.95} & 0.73 & 0.74 & \textbf{0.77} & 0.64 & 0.44 & 0.75 \\
\multicolumn{3}{l}{SpanLLM} & 1.48 & \textbf{1.52} & \textbf{0.87} & \textbf{1.90} & \textbf{2.48} & \textbf{0.62} & \textbf{1.48} & 0.82 & \textbf{0.74} & \textbf{0.90} & 0.75 & \textbf{0.79} & \textbf{0.56} & \textbf{0.80} \\
\bottomrule
    \end{tabular}
    \caption{(\textbf{\textit{Pride and Prejudice}}) Mean Absolute Error (MAE) and Correlation with the gold scores for the predicted component scores averaged across characters and chapters for each method we compare. The initial eight methods apply GPT-4o-mini at different levels of granularity for chapter-level counts. A check in the book column indicates the input chapter text is chunked. A check in the person column indicates the model is asked to label counts for each character individually. A check in the tag column indicates the model is asked about each component individually. The bottom two approaches are our span-level methods. }
    \label{tab:accuracy}
\end{table*}

\section{Results}

First, we assess whether we can accurately label our character components and construct the associated graphs across our two novels for which we have ground truth: \textit{Pride and Prejudice} and \textit{Jane Eyre}.  Then, we use these methods of representation to bring an innovative computational lens to our larger corpus of novels. 

\begin{figure*}[t]
    \centering
    \includegraphics[width=\linewidth]{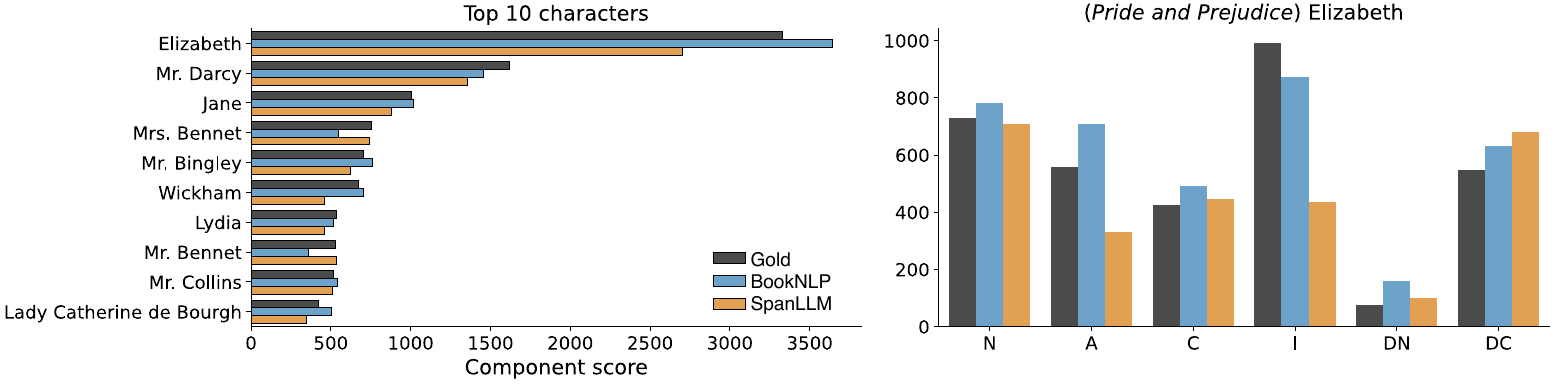}
    \caption{Plots showing component scores from our span-level methods for top characters in \textit{Pride and Prejudice}. The left plot shows total component score for all tags across the top 10 characters and the right plot shows a breakdown of tags by component for the protagonist, Elizabeth.}
    \label{fig:topchars}
\end{figure*}

\begin{figure*}[t]
\centering
\begin{subfigure}[t]{0.35\textwidth}
  \centering
  \includegraphics[width=\linewidth]{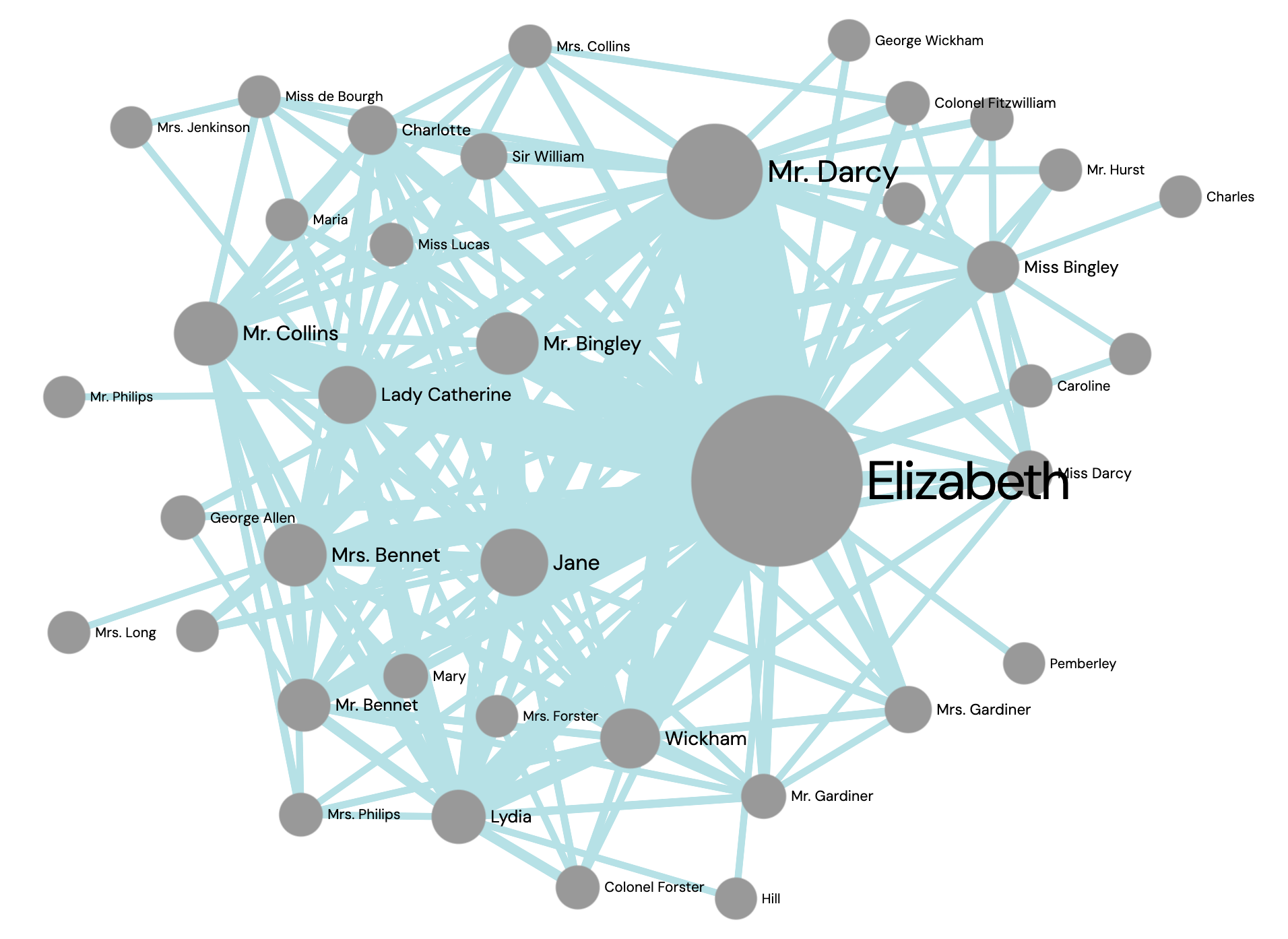}
  \caption{Co-occurrence Network.}
  \label{fig:pnp}
\end{subfigure}
\hfill
\begin{subfigure}[t]{0.26\textwidth}
  \centering
  \includegraphics[width=\linewidth]{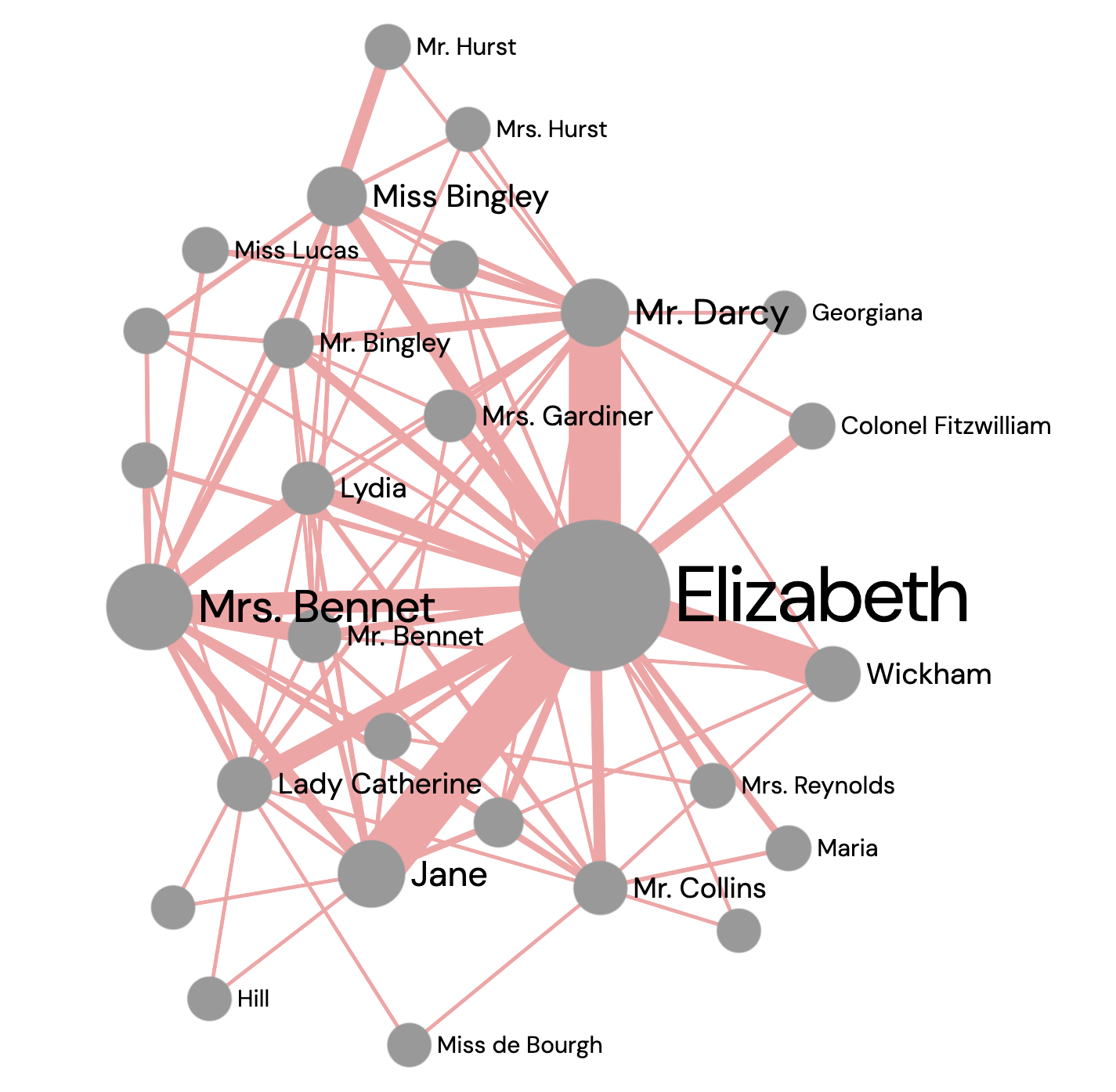}
  \caption{Dialogue Network (C).}
  \label{fig:pnp_dia}
\end{subfigure}
\begin{subfigure}[t]{0.35\textwidth}
  \centering
  \includegraphics[width=\linewidth]{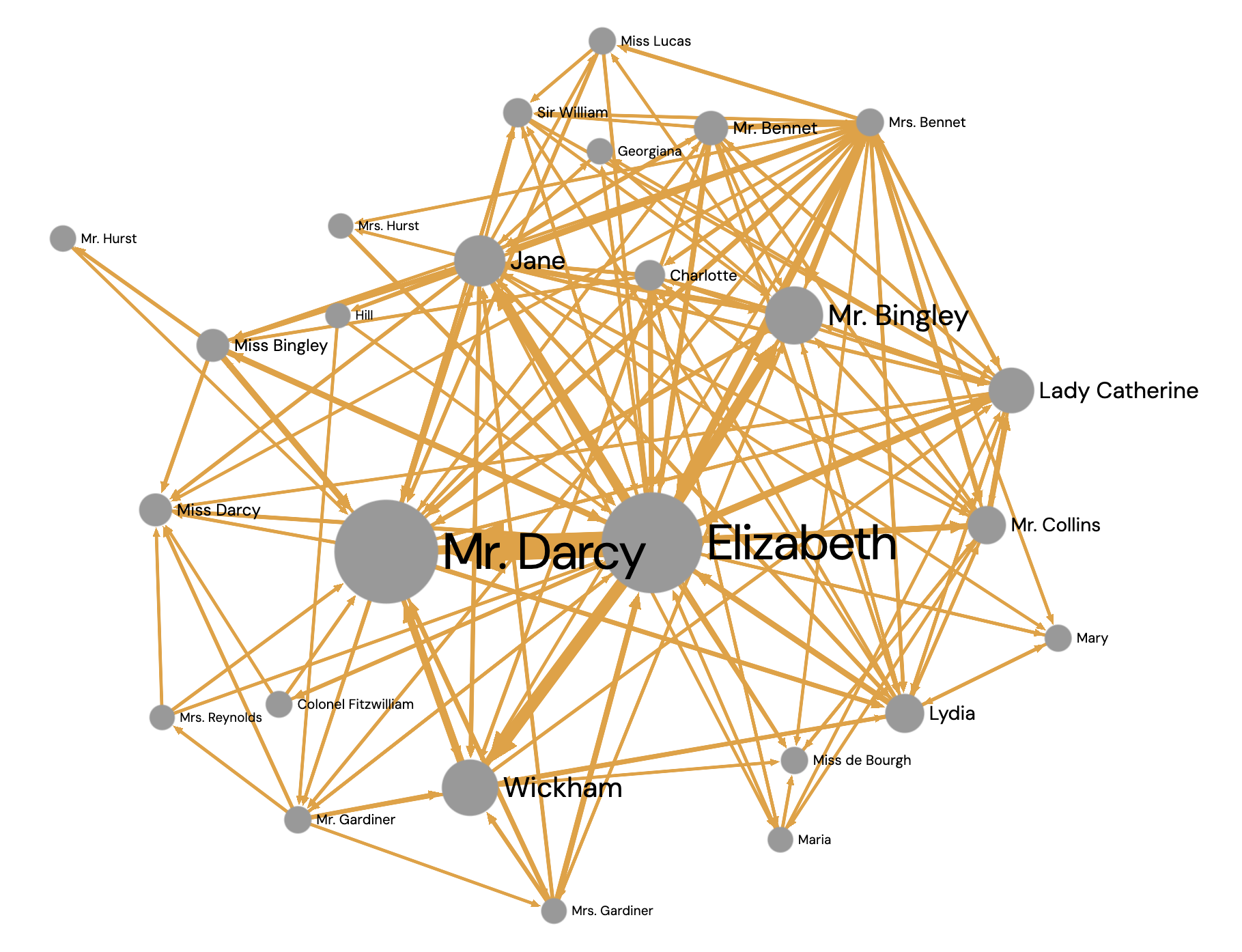}
  \caption{Discussion Network (DC).}
  \label{fig:pnp_gossip}
\end{subfigure}
\caption{Character networks for \textit{Pride \& Prejudice} illustrating the three edge representations chosen.}
\label{fig:pnp}
\end{figure*}

\begin{table}[t]
\centering
\small
\setlength{\tabcolsep}{3pt}
\begin{tabular}{lrr|rr|rr|rr}
\toprule
& \multicolumn{2}{c|}{\textbf{Co-occ.}} &
\multicolumn{2}{c|}{\textbf{Dialogue}} &
\multicolumn{4}{c}{\textbf{Discussion (DC)}} \\
\cmidrule(lr){2-3}
\cmidrule(lr){4-5}
\cmidrule(lr){6-9}
\textbf{Character} &
PR & Rnk. &
PR & Rnk. &
In & Rnk. &
Out & Rnk. \\
\midrule
Elizabeth        & \textbf{0.14} & 1 & \textbf{0.26} & 1 & 311 & 2 & \textbf{524} & 1 \\
Darcy       & 0.07 & 2 & 0.09 & 3 & \textbf{323} & 1 & 237 & 4 \\
Jane             & 0.06 & 3 & 0.07 & 4 & 108 & 5 & 133 & 3 \\
Lady Cath.   & 0.04 & 4 & 0.03 & 11 & 85  & 6 & 64  & 8 \\
Bingley     & 0.04 & 5 & 0.03 & 12 & 136 & 3 & 27  & 14 \\
\bottomrule
\end{tabular}
\caption{
Top characters in \textit{Pride and Prejudice} under different network constructions.
PageRank (PR) is shown for co-occurrence and dialogue networks.
For the discussion (DC) network, in-strength measures how often a character is discussed by others,
while out-strength measures how often a character discusses others.
}
\label{tab:pp-centrality-comparison}
\end{table}
\subsection{Character representations}
\par{\textbf{Component labeling accuracy.}} 
We report the accuracy of different tagging methods for \textit{Pride and Prejudice} in Table \ref{tab:accuracy} and Figure \ref{fig:topchars}. We see that the best scores are consistently achieved by our BookNLP-based and Span-level LLM methods. In using GPT-4o-mini for chapter-level counts (top eight rows of the table), we see that while it reliably outputs a list of tags that correlate with gold counts, it has consistently higher MAE. Interestingly, when prompted with the most granular version of the task (label one component for one character in a short chunk of text), the MAE and correlation actually deteriorates, which we observe is due to substantial overcounting. 

We also see that even though our span-level LLM (SpanLLM) achieves noticable improvements on components for \textit{Pride and Prejudice} by MAE and correlation, it significantly undercounts components A and I for Elizabeth in \textit{Pride and Prejudice} (shown in Figure \ref{fig:topchars}). We also find it performs inconsistently on \textit{Jane Eyre} (shown in Appendix Table \ref{tab:accuracy_je}). Finally, the LLM methods frequently drop minor characters (see plots in Appendix Figure \ref{fig:bottomcounts}). For these reasons, we use our BookNLP-based method for the scaled corpus analysis in the subsequent sections to demonstrate the types of questions that can be answered with this computational lens. We include these LLM results as a caution to researchers looking to use an LLM as a quick labeling tool for qualitative coding. 

Overall, our BookNLP-based pipeline achieves a low average MAE 
of 1.7 and a strong correlation with our gold counts of 0.75. Since the MAE is at the count-level, being off by 1.7 counts on average is quite close to the true count. The bias metrics shown in Appendix \ref{tab:bias_pp} provide additional context on what this margin of error is relative to total count per character per chapter (i.e., this error is more significant if a character has a score of 2 vs. 20 for a given chapter). The MAE and correlation scores reveal the most challenging tags are discussion of characters (DC) and description by narrator (DN), both of which are the most indirect tags. We still see strong correlation in DC though, validating we can move forward with our corpus-level graph-based analysis using this component.

We show more character-specific tag breakdowns in Appendix Figure \ref{fig:allchars} and present an abbreviated version of these results for \textit{Jane Eyre} in Appendix Tables \ref{tab:accuracy_je}, \ref{tab:bias_je}, and Figure \ref{fig:topchars_je}. Without manually correcting the coreference issue shown in Appendix Figure \ref{fig:jane} resulting from a first person narrator, the predicted component scores from our BookNLP pipeline for \text{Jane Eyre} have MAE 4.35 and Correlation 0.44 against the gold scores (compared to MAE 3.00 and Correlation 0.86 with the correction). For this reason, we exclude first-person narratives from the remainder of the analysis and leave their correct resolution to future work. 

\par{\textbf{Graph construction.}} We demonstrate the different methods for graph construction in terms of edge definition in Figure \ref{fig:pnp} for \textit{Pride and Prejudice} (and Appendix Figure \ref{fig:more-jane} for \textit{Jane Eyre}). The co-occurrence network is overwhelmed with edges while the sparser dialogue and discussion networks highlight relationships more clearly. Interestingly, in the dialogue graph (C component), Jane and Elizabeth and Mr. Darcy and Elizabeth have the thickest (and approximately equal) edges of conversation. When we then look at the discussion network graph (DC component) though, we see that Jane and Elizabeth are primarily talking about other people, whereas Darcy and Elizabeth are talking about each other. These insights are illustrated quantitatively in Table \ref{tab:pp-centrality-comparison}, which underscores that ``being talked about" and ``doing the talking" are distinct dimensions of social prominence. These kinds of questions are only answerable through the discussion network which records edges as directed interactions, in contrast to the co-occurrence and dialogue networks, which are undirected graphs.  We report the full corpus averaged measures with additional detail in the Appendix Table \ref{tab:corpus-summary} and novel-level network measures in Appendix Table \ref{tab:corpus-full}. 


\begin{table}[t]
\centering
\small
\begin{tabular}{lc|lcc}
\toprule
\multicolumn{2}{c|}{\textbf{Protagonist Agreement}} & \multicolumn{3}{c}{\textbf{Cross-Network Corr.}} \\
\textit{Comparison} & \textit{Match\%} & \textit{Centrality} & $\rho$ & SD \\
\midrule
N vs PR$_{co}$         & 89.0 & PageRank    & 0.77 & 0.11 \\
N vs EV$_{co}$         & 78.1 & Eigenvector & 0.79 & 0.15 \\
C vs PR$_{di}$         & 84.9 & Degree      & 0.75 & 0.09 \\
C vs EV$_{di}$         & 78.1 & Betweenness & 0.62 & 0.12 \\
N vs C                 & 78.1 &             &      &      \\
PR$_{co}$ vs PR$_{di}$ & 75.3 &             &      &      \\
\bottomrule
\end{tabular}
\caption{Left: Protagonist agreement (\% of novels selecting same top character). Right: Spearman correlation between co-occurrence and dialogue network centrality measures ($n=73$).}
\label{tab:method-comparison}
\end{table}

\begin{table}[t]
\centering
\small
\begin{tabular}{lcccc}
\toprule
\textbf{Tag} &
\textbf{PR$_{co}$} &
\textbf{EV$_{co}$} &
\textbf{PR$_{di}$} &
\textbf{EV$_{di}$} \\
\midrule
N  & \textbf{0.89} & 0.83 & 0.77 & 0.77 \\
C  & 0.66 & 0.61 & \textbf{0.88} & 0.82 \\
I  & 0.69 & 0.64 & 0.77 & 0.75 \\
A  & 0.72 & 0.66 & 0.76 & 0.74 \\
DC & 0.70 & 0.71 & 0.66 & 0.70 \\
DN & 0.63 & 0.58 & 0.68 & 0.66 \\
\bottomrule
\end{tabular}
\caption{
Average Spearman rank correlation between character tags and network centrality measures, computed per novel and averaged across 73 novels. PR = PageRank, EV = Eigenvector centrality; $co$ = co-occurrence network, $di$ = dialogue network.
}
\label{tab:tag-centrality}
\end{table}


\begin{table}[t]
\centering
\small
\begin{tabular}{lccc}
\toprule
\textbf{Measure} & \textbf{Gini} & \textbf{Top-1} & \textbf{Top-1 vs. 2} \\
\midrule
\multicolumn{4}{l}{\emph{Tag-based measures}} \\
N   & 0.77 & 0.23 & 1.87 \\
C   & 0.74 & 0.27 & 1.89 \\
I   & 0.79 & 0.31 & 2.39 \\
A   & 0.80 & 0.29 & 2.26 \\
DC  & 0.70 & 0.21 & 1.65 \\
DN  & 0.69 & 0.29 & 2.39 \\
\midrule
\multicolumn{4}{l}{\emph{Network centrality}} \\
PR$_{co}$ & 0.66 & 0.17 & 1.63 \\
PR$_{di}$ & 0.58 & 0.20 & 1.63 \\
\bottomrule
\end{tabular}
\caption{
Concentration of character importance across novels.
Higher values indicate stronger inequality ($n=73$). Top-1 vs. 2 measures the ratio of centrality between the top two characters.
}
\label{tab:concentration}
\end{table}

\begin{table}[t]
\centering
\small
\setlength{\tabcolsep}{4pt}
\begin{tabular}{lccc}
\toprule
\textbf{Metric} & \textbf{M} & \textbf{F} & \textbf{$p$ (F vs M)} \\
\midrule
In (discussion)   & 0.97$\pm$0.18 & 1.11$\pm$0.28 & .013$^{*}$ \\
Out (discussion)  & 0.89$\pm$0.16 & 1.22$\pm$0.37 & $<\!.001^{***}$ \\
PR           & 0.99$\pm$0.17 & 1.06$\pm$0.28 & .172 \\
\bottomrule
\end{tabular}
\caption{
Top-10 representation ratios in the discussion (DC) network (mean$\pm$SD across novels).
$p$-values are from paired $t$-tests comparing per-novel gender ratios ($n=73$).
$^{*}p<.05$, $^{**}p<.01$, $^{***}p<.001$.
}
\label{tab:dc-gender-rep}
\end{table}

\begin{table}[t]
\centering
\small
\setlength{\tabcolsep}{4pt}
\begin{tabular}{lc}
\toprule
\textbf{Edge type} & \textbf{Share (\%)} \\
\midrule
F$\rightarrow$F (women discussing women) & 16.2 $\pm$ 12.2 \\
F$\rightarrow$M (women discussing men)   & 25.1 $\pm$ 10.9 \\
M$\rightarrow$F (men discussing women)   & 22.5 $\pm$ 10.0 \\
M$\rightarrow$M (men discussing men)     & 36.2 $\pm$ 24.6 \\
\midrule
F$\rightarrow$M / M$\rightarrow$F ratio   & 1.26 $\pm$ 0.72$^{**}$ \\
\bottomrule
\end{tabular}
\caption{
Edge-level gender structure in discussion (DC) networks.
Values show mean$\pm$SD across novels.
The F$\rightarrow$M / M$\rightarrow$F ratio quantifies asymmetric cross-gender attention ($n=73$).
$^{**}p<.01$.
}
\label{tab:dc-edge-gender}
\end{table}

\subsection{Corpus analysis}
In this section, we use our character representations to answer a series of Research Questions (RQs) across our larger corpus of novels. 
\par{\textbf{RQ 1: Is the most significant character consistent across representations?}}
The co-occurrence and dialogue network methods capture overlapping but distinct relational structure in the novels. As such, across the corpus, centrality rankings between these methods exhibit strong correlation (PageRank $\rho = 0.77 \pm 0.11$, eigenvector $\rho = 0.79 \pm 0.15$; Table \ref{tab:method-comparison}). However, the correlation is weaker for betweenness ($\rho = 0.62 \pm 0.12$), suggesting that bridging characters may appear differently in the networks. Protagonist identification is similarly consistent but far from identical: co-occurrence PageRank and dialogue PageRank choose the same top character in only 75.3\% of our novels (Table \ref{tab:method-comparison}).

We exclude the discussion (DC) network from this comparison because it is directed, whereas the co-occurrence and dialogue networks are undirected and readily compared on the basis of degree. However, the DC network reveals additional nuance in protagonist identification; as shown in Table \ref{tab:pp-centrality-comparison}, the top-ranked character differs based on whether we measure in-strength (being discussed) or out-strength (discussing others). In \textit{Pride and Prejudice}, for example, Darcy ranks first by in-strength while Elizabeth ranks first by out-strength, demonstrating how character centrality can be complicated by directed edge dynamics in the discussion network.

We also find that the character component score 
as defined in Section \ref{subsec:component_score} aligns with network centrality in theoretically meaningful ways. We compute correlations between individual component tags and network centrality measures and average across the corpus (Table \ref{tab:tag-centrality}). The N tag (named mentions) correlates most strongly with co-occurrence centrality ($\rho = 0.89$ with co-occurrence PageRank), consistent with co-occurrence capturing narrative presence. Similarly, the C tag (communication) correlates most strongly with dialogue centrality (e.g., $\rho = 0.88$ with dialogue PageRank), consistent with dialogue networks capturing conversational participation. Other components (I/A/DC/DN) show moderate correlations across both networks, suggesting that ``character weight" cannot be reduced in every case to simple speech or co-presence with other characters. 

\par{\textbf{RQ 2: How much space does the central character occupy relative to supporting characters?}}
Narrative attention is unequally distributed between the characters, with the protagonist and core characters accounting for a majority of the ``space.'' 
To quantify this, we compute inequality metrics per novel and average across the corpus (Table \ref{tab:concentration}). The Gini coefficient, as defined in Appendix Table \ref{tab:metrics}, measures the unequal distribution of narrative attention in both the component scores and character networks. A Gini coefficient of 0 indicates perfect equality in narrative attention across characters and 1 indicates perfect inequality (complete concentration of narrative attention in one character). We do not include the discussion network in the network centrality results in this table as it is a directed graph. We also note that the DC tag-based measure corresponds to in-strength in the discussion network and the C tag-based measure is closely related to the out-strength, so these measures can be used to assess the centrality captured by the discussion network. 

In Table \ref{tab:concentration}, we observe the distribution of component scores across characters exhibits high inequality (e.g., Gini 0.69-0.80 across tags). The distribution of node degree is similarly unequal, indicating a high concentration of edge weights associated with the core characters (PageRank Gini 0.66 in co-occurrence; 0.58 in dialogue). 
The pattern of narrative attention described by the degree distribution of the networks, as well as by the component scores, is highly uneven and assigns disproportionate weight to a small cast of core characters.  
Additionally, the less than perfect protagonist agreement in Table \ref{tab:method-comparison} seems to suggest a degree of uncertainty as to who exactly the ``one" is. In relation to Woloch's ``one vs. the many," the ``one" is not one; instead, we find a centralized cluster of several characters, with multiple degrees of "many" constituting the periphery. 

\par{\textbf{RQ 3: What role does gender play in how characters center each other in discussion?}}
When a character discusses another character, they are giving them space in the narrative. We can, therefore, use our discussion network graph to analyze whether gender plays a role in how characters are centered through discussion. 
At the node level, we observe that female characters are overrepresented among the most central discussants of other characters: in the Top-10 by out-strength (``discussing"), female characters had significantly higher representation ratios than males ($1.22 \pm 0.37$ vs. $0.89 \pm 0.16, p <.001; \text{Table } \ref{tab:dc-gender-rep}$). Females are also overrepresented among Top-10 by in-strength (``discussed") ($1.11 \pm 0.28, p=.013 \text{ vs. males})$, while PageRank exhibits no significant gender difference.

Edge-level analysis of the DC graphs further shows asymmetric cross-gender attention, with edges from women to men occurring more frequently than from men to women ($F\rightarrow M/M\rightarrow F = 1.26 \pm 0.72, p=0.003$; Table \ref{tab:dc-edge-gender}). Together, these results suggest that discussion networks capture a distinct social layer: who is positioned as a discussant and who is the topic of discussion, revealing gendered structure not discernible in the undirected co-occurrence or dialogue networks.



\section{Conclusion}
We introduce a new framing and dataset for studying the components of character in classic literature. We evaluate and release different methods of labeling these components in novels to facilitate future work. Finally, we use this computational lens to present a different perspective on character prominence than Woloch's classic theory of ``The One vs. the Many," and to analyze the role gender plays in who gets discussed the most in character dialogue. In the future, we hope to improve these methods for first-person narratives as well. 

While we have focused heavily on the dynamics of discussion through the DC network as well as mention and dialogue-based notions of centrality, future work may extend analysis of the interiority (I) and narrator description (DN) components. Additionally, while we process and analyze character at the novel-level, future work might study narrative dynamics as change in character prominence over time at the chapter level or across multi-volume works. 

Whereas previous studies focused on either character networks or character attribute detection, we integrate attribute detection with a new discussion-based network, adding a layer of complexity to prior accounts of literary character significance.

\section{Limitations}
We acknowledge limitations of this work including the chosen corpus of texts, which is restricted to third-person, 19th-century English-language novels only. This potentially limits the applicability of our methods and results to different cultural settings, historical periods, and narrative styles. We also note the limitations of our computational methods with respect to coreference resolution for first-person narrators more generally, which affects the ability to construct accurate and scalable network analyses of characters in complex novels. LLMs in our work also exhibited difficulty in reproducing the tagging scheme in a computationally efficient and reliable way that could be transferred across a large set of novels without significant context engineering. Finally, due to the high manual cost of annotating a novel according to the literary theory of character described, we are limited to two novels as sources of ground truth, which may limit the generality of our methods.
\bibliography{custom}

@book{schaffer2021communities,
  title={Communities of care: The social ethics of {Victorian} fiction},
  author={Schaffer, Talia},
  year={2021},
  publisher={Princeton University Press},
  address   = {Princeton, NJ}
}

@book{Woloch2003,
  author    = {Alex Woloch},
  title     = {The One vs.\ the Many: Minor Characters and the Space of the Protagonist in the Novel},
  publisher = {Princeton University Press},
  address   = {Princeton, NJ},
  year      = {2003},
  isbn      = {0691113130, 0691113149},
}

@book{Propp1968,
  author    = {Vladimir I. Propp},
  title     = {Morphology of the Folktale},
  translator = {Laurence Scott},
  editor    = {Louis A. Wagner},
  publisher = {University of Texas Press},
  address   = {Austin, TX},
  year      = {1968},
  series    = {Publications of the American Folklore Society, Bibliographical and Special Series, Vol.\ 9},
  isbn      = {9780292783768, 0292783760},
  note      = {English translation of *Morfologiya skazki* (1928)},
}

@book{Figlerowicz2016,
  author    = {Marta Figlerowicz},
  title     = {Flat Protagonists: A Theory of Novel Character},
  publisher = {Oxford University Press},
  address   = {Oxford \& New York},
  year      = {2016},
  isbn      = {9780190496760, 9780190496777},
}

@book{Forster1927,
  author    = {E. M. Forster},
  title     = {Aspects of the Novel},
  publisher = {Edward Arnold},
  address   = {London},
  year      = {1927},
}

@inproceedings{elson-etal-2010-extracting,
    title = "Extracting Social Networks from Literary Fiction",
    author = "Elson, David  and
      Dames, Nicholas  and
      McKeown, Kathleen",
    editor = "Haji{\v{c}}, Jan  and
      Carberry, Sandra  and
      Clark, Stephen  and
      Nivre, Joakim",
    booktitle = "Proceedings of the 48th Annual Meeting of the Association for Computational Linguistics",
    month = jul,
    year = "2010",
    address = "Uppsala, Sweden",
    publisher = "Association for Computational Linguistics",
    url = "https://aclanthology.org/P10-1015/",
    pages = "138--147"
}

@inproceedings{piper-etal-2021-narrative,
    title = "Narrative Theory for Computational Narrative Understanding",
    author = "Piper, Andrew  and
      So, Richard Jean  and
      Bamman, David",
    editor = "Moens, Marie-Francine  and
      Huang, Xuanjing  and
      Specia, Lucia  and
      Yih, Scott Wen-tau",
    booktitle = "Proceedings of the 2021 Conference on Empirical Methods in Natural Language Processing",
    month = nov,
    year = "2021",
    address = "Online and Punta Cana, Dominican Republic",
    publisher = "Association for Computational Linguistics",
    url = "https://aclanthology.org/2021.emnlp-main.26/",
    doi = "10.18653/v1/2021.emnlp-main.26",
    pages = "298--311"
}

@inproceedings{thai-iyyer-2025-literary,
    title = "Literary Evidence Retrieval via Long-Context Language Models",
    author = "Thai, Katherine  and
      Iyyer, Mohit",
    editor = "Che, Wanxiang  and
      Nabende, Joyce  and
      Shutova, Ekaterina  and
      Pilehvar, Mohammad Taher",
    booktitle = "Proceedings of the 63rd Annual Meeting of the Association for Computational Linguistics (Volume 2: Short Papers)",
    month = jul,
    year = "2025",
    address = "Vienna, Austria",
    publisher = "Association for Computational Linguistics",
    url = "https://aclanthology.org/2025.acl-short.29/",
    doi = "10.18653/v1/2025.acl-short.29",
    pages = "369--380",
    ISBN = "979-8-89176-252-7"
}

@inproceedings{amalvy-etal-2025-role,
    title = "The Role of Natural Language Processing Tasks in Automatic Literary Character Network Construction",
    author = "Amalvy, Arthur  and
      Labatut, Vincent  and
      Dufour, Richard",
    editor = "Rambow, Owen  and
      Wanner, Leo  and
      Apidianaki, Marianna  and
      Al-Khalifa, Hend  and
      Eugenio, Barbara Di  and
      Schockaert, Steven",
    booktitle = "Proceedings of the 31st International Conference on Computational Linguistics",
    month = jan,
    year = "2025",
    address = "Abu Dhabi, UAE",
    publisher = "Association for Computational Linguistics",
    url = "https://aclanthology.org/2025.coling-main.566/",
    pages = "8462--8473"
}

@inproceedings{soni-etal-2023-grounding,
    title = "Grounding Characters and Places in Narrative Text",
    author = "Soni, Sandeep  and
      Sihra, Amanpreet  and
      Evans, Elizabeth  and
      Wilkens, Matthew  and
      Bamman, David",
    editor = "Rogers, Anna  and
      Boyd-Graber, Jordan  and
      Okazaki, Naoaki",
    booktitle = "Proceedings of the 61st Annual Meeting of the Association for Computational Linguistics (Volume 1: Long Papers)",
    month = jul,
    year = "2023",
    address = "Toronto, Canada",
    publisher = "Association for Computational Linguistics",
    url = "https://aclanthology.org/2023.acl-long.655/",
    doi = "10.18653/v1/2023.acl-long.655",
    pages = "11723--11736"
}

@inproceedings{bamman-etal-2020-annotated,
    title = "An Annotated Dataset of Coreference in {E}nglish Literature",
    author = "Bamman, David  and
      Lewke, Olivia  and
      Mansoor, Anya",
    editor = "Calzolari, Nicoletta  and
      B{\'e}chet, Fr{\'e}d{\'e}ric  and
      Blache, Philippe  and
      Choukri, Khalid  and
      Cieri, Christopher  and
      Declerck, Thierry  and
      Goggi, Sara  and
      Isahara, Hitoshi  and
      Maegaard, Bente  and
      Mariani, Joseph  and
      Mazo, H{\'e}l{\`e}ne  and
      Moreno, Asuncion  and
      Odijk, Jan  and
      Piperidis, Stelios",
    booktitle = "Proceedings of the Twelfth Language Resources and Evaluation Conference",
    month = may,
    year = "2020",
    address = "Marseille, France",
    publisher = "European Language Resources Association",
    url = "https://aclanthology.org/2020.lrec-1.6/",
    pages = "44--54",
    language = "eng",
    ISBN = "979-10-95546-34-4"
}

@article{labatut2019extraction,
  title={Extraction and analysis of fictional character networks: A survey},
  author={Labatut, Vincent and Bost, Xavier},
  journal={ACM Computing Surveys (CSUR)},
  volume={52},
  number={5},
  pages={1--40},
  year={2019},
  publisher={ACM New York, NY, USA}
}

@article{moretti2011network,
  title={Network theory, plot analysis},
  author={Moretti, Franco},
  year={2011},
  publisher={Deutsche Nationalbibliothek}
}

@article{beveridge2016network,
  title={Network of thrones},
  author={Beveridge, Andrew and Shan, Jie},
  journal={Math Horizons},
  volume={23},
  number={4},
  pages={18--22},
  year={2016},
  publisher={Taylor \& Francis}
}

@inproceedings{bonato2016mining,
  title={Mining and modeling character networks},
  author={Bonato, Anthony and D’Angelo, David Ryan and Elenberg, Ethan R and Gleich, David F and Hou, Yangyang},
  booktitle={International workshop on algorithms and models for the web-graph},
  pages={100--114},
  year={2016},
  organization={Springer}
}

@inproceedings{agarwal-etal-2012-social,
    title = "Social Network Analysis of \textit{Alice in Wonderland}",
    author = "Agarwal, Apoorv  and
      Corvalan, Augusto  and
      Jensen, Jacob  and
      Rambow, Owen",
    editor = "Elson, David  and
      Kazantseva, Anna  and
      Mihalcea, Rada  and
      Szpakowicz, Stan",
    booktitle = "Proceedings of the {NAACL}-{HLT} 2012 Workshop on Computational Linguistics for Literature",
    month = jun,
    year = "2012",
    address = "Montr{\'e}al, Canada",
    publisher = "Association for Computational Linguistics",
    url = "https://aclanthology.org/W12-2513/",
    pages = "88--96"
}

@inproceedings{li2019complex,
  title={Complex networks of characters in fictional novels},
  author={Li, Jiarong and Zhang, Chi and Tan, Huan and Li, Chunfang},
  booktitle={2019 IEEE/ACIS 18th International Conference on Computer and Information Science (ICIS)},
  pages={417--420},
  year={2019},
  organization={IEEE}
}

@article{masias2017exploring,
  title={Exploring the prominence of \textit{Romeo and Juliet}’s characters using weighted centrality measures},
  author={Mas{\'\i}as, V{\'\i}ctor Hugo and Baldwin, Paula and Laengle, Sigifredo and Vargas, Augusto and Crespo, Fernando A},
  journal={Digital Scholarship in the Humanities},
  volume={32},
  number={4},
  pages={837--858},
  year={2017},
  publisher={Oxford University Press}
}

@inproceedings{evalyn2018analyzing,
  title={Analyzing Social Networks of XML Plays: Exploring Shakespeare's Genres.},
  author={Evalyn, Lawrence and Gauch, Susan and Shukla, Manisha},
  booktitle={DH},
  pages={368--370},
  year={2018}
}

@inproceedings{hamilton-etal-2025-city,
    title = "A City of Millions: Mapping Literary Social Networks At Scale",
    author = "Hamilton, Sil  and
      Hicke, Rebecca  and
      Mimno, David  and
      Wilkens, Matthew",
    editor = {H{\"a}m{\"a}l{\"a}inen, Mika  and
      {\"O}hman, Emily  and
      Bizzoni, Yuri  and
      Miyagawa, So  and
      Alnajjar, Khalid},
    booktitle = "Proceedings of the 5th International Conference on Natural Language Processing for Digital Humanities",
    month = may,
    year = "2025",
    address = "Albuquerque, USA",
    publisher = "Association for Computational Linguistics",
    url = "https://aclanthology.org/2025.nlp4dh-1.46/",
    doi = "10.18653/v1/2025.nlp4dh-1.46",
    pages = "543--549",
    ISBN = "979-8-89176-234-3"
}

@techreport{hagberg2007exploring,
  title={Exploring network structure, dynamics, and function using NetworkX},
  author={Hagberg, Aric and Swart, Pieter J and Schult, Daniel A},
  year={2007},
  institution={Los Alamos National Laboratory (LANL)}
}

@inproceedings{bastian2009gephi,
  title={Gephi: an open source software for exploring and manipulating networks},
  author={Bastian, Mathieu and Heymann, Sebastien and Jacomy, Mathieu},
  booktitle={Proceedings of the international AAAI conference on web and social media},
  volume={3},
  number={1},
  pages={361--362},
  year={2009}
}

@article{nickel2017poincare,
  title={Poincar{\'e} embeddings for learning hierarchical representations},
  author={Nickel, Maximillian and Kiela, Douwe},
  journal={Advances in neural information processing systems},
  volume={30},
  year={2017}
}

@article{menon2019keeping,
  title={Keeping Count},
  author={Menon, Tara},
  journal={Narrative},
  volume={27},
  number={2},
  pages={160--181},
  year={2019},
  publisher={JSTOR}
}

@inproceedings{rashkin-etal-2018-modeling,
    title = "Modeling Naive Psychology of Characters in Simple Commonsense Stories",
    author = "Rashkin, Hannah  and
      Bosselut, Antoine  and
      Sap, Maarten  and
      Knight, Kevin  and
      Choi, Yejin",
    editor = "Gurevych, Iryna  and
      Miyao, Yusuke",
    booktitle = "Proceedings of the 56th Annual Meeting of the Association for Computational Linguistics (Volume 1: Long Papers)",
    month = jul,
    year = "2018",
    address = "Melbourne, Australia",
    publisher = "Association for Computational Linguistics",
    url = "https://aclanthology.org/P18-1213/",
    doi = "10.18653/v1/P18-1213",
    pages = "2289--2299"
}

@phdthesis{schuler2005verbnet,
  title={VerbNet: A broad-coverage, comprehensive verb lexicon},
  author={Schuler, Karin Kipper},
  year={2005},
  school={University of Pennsylvania}
}

@inproceedings{bamman-etal-2019-annotated,
    title = "An annotated dataset of literary entities",
    author = "Bamman, David  and
      Popat, Sejal  and
      Shen, Sheng",
    editor = "Burstein, Jill  and
      Doran, Christy  and
      Solorio, Thamar",
    booktitle = "Proceedings of the 2019 Conference of the North {A}merican Chapter of the Association for Computational Linguistics: Human Language Technologies, Volume 1 (Long and Short Papers)",
    month = jun,
    year = "2019",
    address = "Minneapolis, Minnesota",
    publisher = "Association for Computational Linguistics",
    url = "https://aclanthology.org/N19-1220/",
    doi = "10.18653/v1/N19-1220",
    pages = "2138--2144"
}

@article{piper2024characters,
  title={What do characters do? The embodied agency of fictional characters},
  author={Piper, Andrew},
  journal={Journal of Computational Literary Studies},
  volume={2},
  number={1},
  year={2024},
  publisher={Universit{\"a}ts-und Landesbibliothek Darmstadt}
}

@article{margolin2002naming,
  title={Naming and believing: practices of the proper name in narrative fiction},
  author={Margolin, Uri},
  journal={Narrative},
  volume={10},
  number={2},
  pages={107--127},
  year={2002},
  publisher={JSTOR}
}

@book{barthes1974s,
  title={S/Z: an essay},
  author={Barthes, Roland},
  year={1974},
  publisher={Macmillan}
}

@book{eder2010characters,
  title={Characters in fictional worlds: Understanding imaginary beings in literature, film, and other media},
  author={Eder, Jens and Jannidis, Fotis and Schneider, Ralf},
  volume={3},
  year={2010},
  publisher={Walter de Gruyter}
}

@article{menon2024forms,
  title={Forms of Discourse: The Significance of Direct Speech},
  author={Menon, Tara K},
  journal={Studies in the Novel},
  volume={56},
  number={4},
  pages={376--387},
  year={2024},
  publisher={Johns Hopkins University Press}
}

@inproceedings{menon2024constructing,
  title={Constructing Attachment: Persistent and Elided Speech in {Jane Austen}'s Novels},
  author={Menon, Tara K},
  booktitle={Novel: A Forum on Fiction},
  volume={57},
  number={3},
  pages={375--398},
  year={2024},
  organization={Duke University Press}
}

@inproceedings{sims-bamman-2020-measuring,
    title = "Measuring Information Propagation in Literary Social Networks",
    author = "Sims, Matthew  and
      Bamman, David",
    editor = "Webber, Bonnie  and
      Cohn, Trevor  and
      He, Yulan  and
      Liu, Yang",
    booktitle = "Proceedings of the 2020 Conference on Empirical Methods in Natural Language Processing (EMNLP)",
    month = nov,
    year = "2020",
    address = "Online",
    publisher = "Association for Computational Linguistics",
    url = "https://aclanthology.org/2020.emnlp-main.47/",
    doi = "10.18653/v1/2020.emnlp-main.47",
    pages = "642--652"
}

@inproceedings{rehurek_lrec,
      title = {{Software Framework for Topic Modelling with Large Corpora}},
      author = {Radim {\v R}eh{\r u}{\v r}ek and Petr Sojka},
      booktitle = {{Proceedings of the LREC 2010 Workshop on New
           Challenges for NLP Frameworks}},
      pages = {45--50},
      year = 2010,
      month = May,
      day = 22,
      publisher = {ELRA},
      address = {Valletta, Malta},
      note={\url{http://is.muni.cz/publication/884893/en}},
      language={English}
}

@inproceedings{bamman-etal-2014-bayesian,
    title = "A {B}ayesian Mixed Effects Model of Literary Character",
    author = "Bamman, David  and
      Underwood, Ted  and
      Smith, Noah A.",
    editor = "Toutanova, Kristina  and
      Wu, Hua",
    booktitle = "Proceedings of the 52nd Annual Meeting of the Association for Computational Linguistics (Volume 1: Long Papers)",
    month = jun,
    year = "2014",
    address = "Baltimore, Maryland",
    publisher = "Association for Computational Linguistics",
    url = "https://aclanthology.org/P14-1035/",
    doi = "10.3115/v1/P14-1035",
    pages = "370--379"
}

@article{barthes1968reality,
  title={The reality effect},
  author={Barthes, Roland},
  journal={The Novel: An Anthology of Criticism and Theory, 1900-2000},
  pages={229--34},
  year={1968}
}

@article{gerlach2020standardized,
  title={A standardized {Project Gutenberg} corpus for statistical analysis of natural language and quantitative linguistics},
  author={Gerlach, Martin and Font-Clos, Francesc},
  journal={Entropy},
  volume={22},
  number={1},
  pages={126},
  year={2020},
  publisher={MDPI}
}

@inproceedings{piper-etal-2024-social,
    title = "The Social Lives of Literary Characters: Combining citizen science and language models to understand narrative social networks",
    author = "Piper, Andrew  and
      Xu, Michael  and
      Ruths, Derek",
    editor = {H{\"a}m{\"a}l{\"a}inen, Mika  and
      {\"O}hman, Emily  and
      Miyagawa, So  and
      Alnajjar, Khalid  and
      Bizzoni, Yuri},
    booktitle = "Proceedings of the 4th International Conference on Natural Language Processing for Digital Humanities",
    month = nov,
    year = "2024",
    address = "Miami, USA",
    publisher = "Association for Computational Linguistics",
    url = "https://aclanthology.org/2024.nlp4dh-1.45/",
    doi = "10.18653/v1/2024.nlp4dh-1.45",
    pages = "472--482"
}

@misc{Marcus2025CaringCharacter,
  author       = {Marcus, Sharon},
  title        = {Caring about Character},
  howpublished = {Keynote lecture, Victorians Institute Annual Conference},
  address      = {Greenville, South Carolina},
  institution  = {Furman University},
  month        = sep,
  year         = {2025},
  note         = {Hosted by Furman University}
}

\appendix
\onecolumn

\section{Additional metric definitions}
We define bias as,
\begin{equation*}
\text{Bias} = \frac{\sum_{c\in C, k\in K}p_{ck} - \sum_{c\in C, k\in K}g_{ck}}{\sum_{c\in C, k\in K}g_{ck}}
\end{equation*}
The results for bias are shown in Tables \ref{tab:bias_pp} and \ref{tab:bias_je}.

\begin{table}[h]
\centering \small
    \begin{tabular}{ccc|cccccc|c}
    \toprule
    \multicolumn{3}{c|}{Method}& \multicolumn{7}{c}{Bias}\\\midrule
        \tiny \faBook&\tiny\faUser&\tiny\faTag  & N & A & C & I & DC & DN & avg. \\\midrule\midrule
&& & -0.2 & -0.55 & 0.05 & -0.38 & -0.66 & 2.52 & -0.44 \\
&&\tiny\checkmark & -0.11 & 1.08 & 1.24 & 0.11 & -0.48 & 4.56 & -0.05 \\
&\tiny\checkmark& & 0.56 & 0.02 & 1.14 & 0.14 & -0.29 & 9.22 & 0.17 \\
&\tiny\checkmark&\tiny\checkmark & 0.77 & 3.53 & 3.26 & 5.86 & 0.18 & 9.12 & 1.65 \\
\tiny\checkmark&& & 0.65 & -0.46 & 0.34 & 0.1 & -0.46 & 4.71 & 0.01 \\
\tiny\checkmark&&\tiny\checkmark & 0.5 & 2.97 & 2.67 & 1.43 & 0.04 & 8.44 & 0.79 \\
\tiny\checkmark&\tiny\checkmark& & 2.14 & 0.2 & 1.61 & 1.62 & 0.08 & 15.52 & 1.17 \\
\tiny\checkmark&& & 1.09 & 8.72 & 3.78 & 13.56 & 1.14 & 14.74 & 3.74 \\
\midrule
\multicolumn{3}{c|}{BookNLP} & 0.06 & 0.76 & 0.17 & 0.3 & -0.02 & 0.43 & 0.03 \\
\multicolumn{3}{c|}{SpanLLM} & -0.11 & -0.05 & 0.29 & -0.27 & -0.06 & 0.84 & -0.12 \\
\bottomrule
    \end{tabular}
    \caption{Bias scores for \textit{Pride and Prejudice} for the predicted component counts averaged across characters and chapters for each method we compare. The initial eight methods apply GPT-4o-mini at different levels of granularity for chapter-level counts. A check in the book column indicates the input chapter text is chunked. A check in the person column indicates the model is asked to tag each character individually. A check in the tag column indicates the model is asked about each component individually. The bottom two approaches are our span-level method.}
    \label{tab:bias_pp}
\end{table}

\begin{table}[h]
\centering \small
    \begin{tabular}{cc|cccccc|c}
    \toprule
   \multicolumn{2}{c|}{Method} & \multicolumn{7}{c}{Bias}\\\midrule
         \faBook & \faTag & N & A & C & I & DC & DN & avg. \\\midrule\midrule

 &  & -0.30 & -0.90 & -0.76 & -0.77 & -0.74 & 0.62 & -0.73 \\
 & \tiny\checkmark & -0.32 & -0.62 & -0.47 & -0.44 & -0.68 & 0.84 & -0.54 \\
\tiny\checkmark &  & 0.95 & -0.73 & -0.44 & -0.42 & -0.57 & 1.87 & -0.34 \\
\tiny\checkmark & \tiny\checkmark & 0.55 & 0.18 & 0.19 & 0.60 & -0.06 & 3.66 & 0.26 \\
\midrule
\multicolumn{2}{c|}{BookNLP} & -0.01 & -0.64 & -0.53 & -0.74 & -0.20 & -0.75 & -0.51 \\\midrule
\multicolumn{2}{c|}{BookNLP \faHandPaper} & -0.01 & -0.22 & -0.24 & -0.18 & 0.00 & -0.26 & -0.20 \\
\multicolumn{2}{c|}{SpanLLM} & -0.37 & -0.76 & 0.14 & -0.80 & 0.13 & -0.45 & -0.44 \\
\midrule
\bottomrule
    \end{tabular}
    \caption{Bias scores for \textit{Jane Eyre} for the predicted component counts averaged across characters and chapters for each method we compare. The initial four methods apply GPT-4o-mini at different levels of granularity for chapter-level counts. A check in the book column indicates the input chapter text is chunked. A check in the tag column indicates the model is asked about each component individually. The bottom two approaches are our span-level methods. The BookNLP method with the end indicates we manually intervened to correct the co-reference issue with one of Jane's mention clusters. The BookNLP score without this intervention is also shown.}
    \label{tab:bias_je}
\end{table}

Centrality measures are defined in Table \ref{tab:metrics}.

\begin{table*}
\centering

\small
\begin{tabular}{@{}l p{4cm} p{9cm}@{}}
\toprule
\textbf{Measure} & \textbf{Formula} & \textbf{Interpretation} \\
\midrule
\multicolumn{3}{@{}l@{}}{\textit{Centrality Measures (per character)}}\\[2pt]
\midrule
Degree & $k_v = |\{u : (v,u) \in E\}|$ & Number of unique interaction partners; basic social connectivity\\[4pt]
Strength & $s_v = \sum_{u \in N(v)} w_{vu}$ & Total interaction volume (weighted degree) \\[4pt]
Betweenness & $C_B(v) = \sum_{s \neq v \neq t} \frac{\sigma_{st}(v)}{\sigma_{st}}$ & Fraction of shortest paths through $v$; bridging role \\[4pt]
Closeness & $C_C(v) = \frac{n-1}{\sum_u d(v,u)}$ & Inverse average distance to others\\[4pt]
Eigenvector & $\mathbf{Ax} = \lambda_1 \mathbf{x}$ & Centrality weighted by neighbors' centrality \\[4pt]
PageRank & $\pi = d\mathbf{P}^\top\pi + \frac{1-d}{n}\mathbf{1}$ & Random walk stationary distribution; iterative prestige \\[4pt]
\midrule
\multicolumn{3}{@{}l@{}}{\textit{Global Structure (per network)}}\\[2pt]
\midrule
Gini coefficient & $G = \frac{\sum_i \sum_j |k_i - k_j|}{2n \sum_i k_i}$ & Degree inequality; concentration of connectivity\\[4pt]
Centralization & $C_D = \frac{\sum_v (k_{\max} - k_v)}{(n-1)(n-2)}$ & Degree concentration; 1 for a star network\\[4pt]
Assortativity & $r = \text{corr}(k_u, k_v)_{(u,v) \in E}$ & Pearson correlation of endpoint degrees across edges\\[4pt]
Clustering & $\bar{C} = \frac{1}{n}\sum_v \frac{|\text{triangles at } v|}{\binom{k_v}{2}}$ & Triangle density; neighborhood cohesion\\[4pt]
$\delta$-Hyperbolicity & $\delta = \max_{a,b,c,d} \frac{1}{2}(S_2 - S_1)$ & Tree-likeness; lower indicates more hierarchical structure\\[4pt]
Reciprocity &
$r = \frac{|\{(i,j) \in E : (j,i) \in E\}|}{|E|}$ &
Fraction of directed edges that are mutual; interaction symmetry\\[4pt]
\bottomrule
\end{tabular}
\begin{flushleft}
\footnotesize
\textit{Note:} $w_{vu}$ = edge weight (co-occurrence count), $N(v)$ = neighbors of $v$, $\sigma_{st}$ = number of shortest paths from $s$ to $t$, $\mathbf{A}$ = weighted adjacency matrix, $\mathbf{P}$ = transition matrix, $d$ = damping factor. Betweenness and closeness use distance $= 1/w$ so stronger ties yield shorter paths. For the four-point condition, $S_1 \leq S_2 \leq S_3$ are the sorted sums of opposite pairwise distances.
\end{flushleft}
\caption{Summary of Network Measures: Centrality \& Global Structure}
\label{tab:metrics}
\end{table*}

We compute network centrality measures and create graphs using NetworkX \citep{hagberg2007exploring}, a Python library for the construction, visualization and analysis of graphs and networks. And compute statistics for centrality and global structure across the corpus of 64 novels. The outputs are stored in the GraphML file format for reproducibility and are visualized using Gephi, an open-source application for visualizing large networks and graphs \citep{bastian2009gephi}. We report the average across our corpus for the global measures, and take a more fine-grained view with respect to the centrality measures, which are computed at the node level for the top characters according to a baseline degree ranking. 

We additionally assess the efficiency of our LLM tagging pipelines. We report the input token ratio, output token ratio, and relative runtime. The Input Token Ratio (\textbf{Input TR}) is computed as the total count of input tokens for the prompts required to process a chapter scaled by the token-length of the chapter. The Output Token Ratio (\textbf{Output TR}) is the same for the generated responses to prompts. The Relative Elapsed Time (\textbf{Relative ET}) is the total runtime of prompt calls for a chapter scaled by the token-length of the chapter (and multiplied by 100 to magnify effects). Intuitively, the input/output token ratios indicate a cost multiplier for cost of API calls relative to how long your document is. And the time multiplier gives a runtime estimate based on document length. These results are shown in Table \ref{tab:cost}.

\begin{table}
\centering \small
    \begin{tabular}{ccc|ccc}
    \toprule
         \multicolumn{3}{c|}{Method} &&&\\
         \faBook&\faUser&\faTag & Input TR & Output TR & RE Time \\\midrule
&& & 1.16 & 0.12 & 0.25 \\
&&\checkmark & 6.58 & 0.16 & 0.46 \\
&\checkmark& & 14.44 & 0.27 & 0.92 \\
&\checkmark&\checkmark & 78.92 & 0.43 & 3.76 \\
\checkmark&& & 2.85 & 0.67 & 1.5 \\
\checkmark&&\checkmark & 12.69 & 0.79 & 2.83 \\
\checkmark&\checkmark& & 30.54 & 2.68 & 7.84 \\
\checkmark&\checkmark&\checkmark & 134.79 & 4.07 & 24.89 \\
\midrule
\multicolumn{3}{c|}{SpanLLM} & 44.43 & 1.42 & 4.5 \\
\bottomrule
    \end{tabular}
    \caption{The Input Token Ratio (TR) indicates the average input tokens required for prompts per chapter normalized by the length of that chapter. The Output Token Ratio (TR) is the same for output tokens. The relative time is the runtime for the prompts required normalized by chapter length (and multiplied by 100 to show effects). The initial eight methods apply GPT-4o-mini at different levels of granularity for chapter-level counts. A check in the book column indicates the input chapter text is chunked. A check in the person column indicates the model is asked to tag each character individually. A check in the tag column indicates the model is asked about each component individually. The bottom approach is our span-level methods.}
    \label{tab:cost}
\end{table}

\newpage
\section{Poincar\'{e} embeddings}
\label{sec:poin}
\begin{figure}[h]
    \centering
    \includegraphics[width=0.6\linewidth]{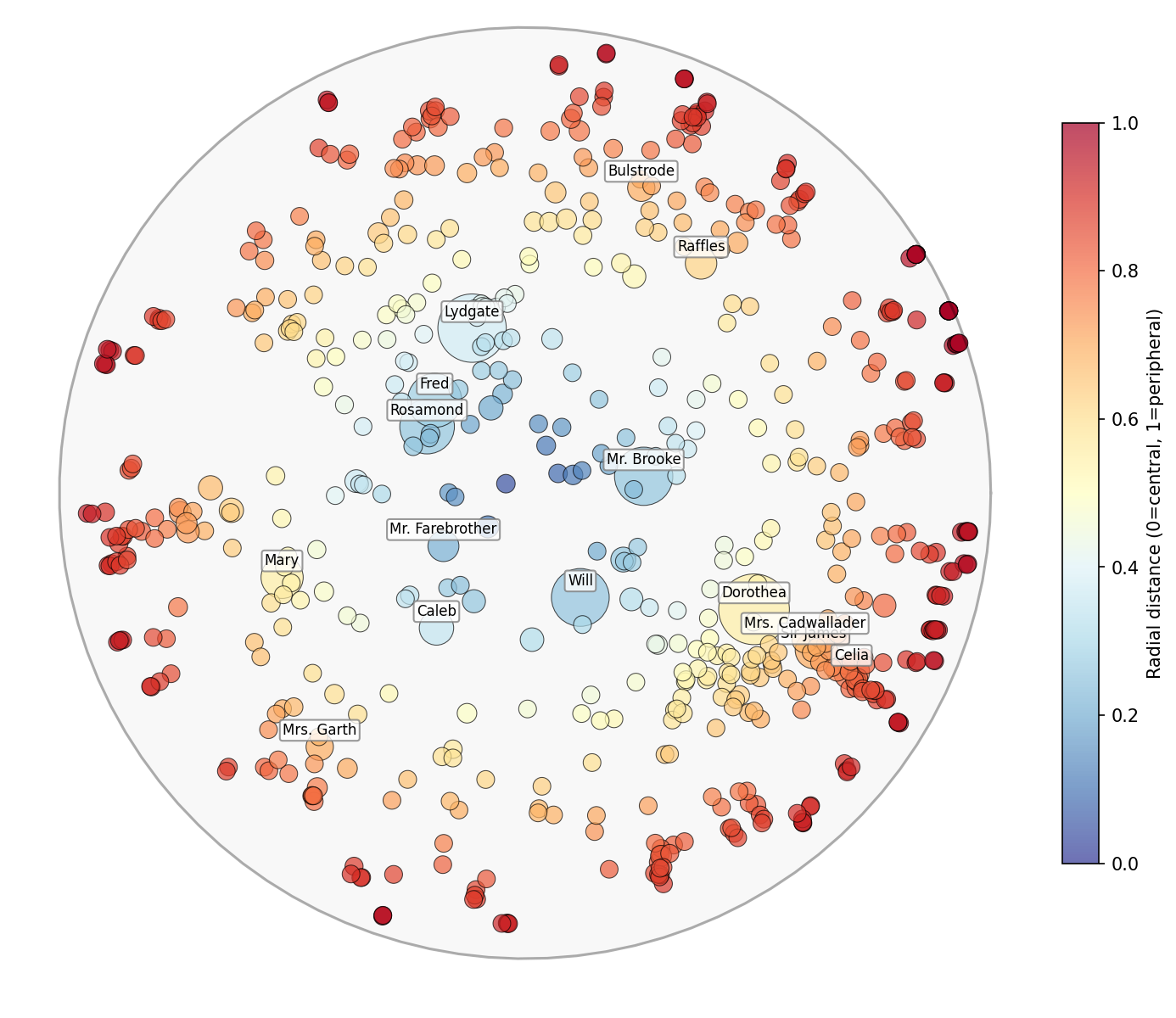}
    \caption{Poincar\'{e} disk embedding of the \textit{Middlemarch} co-occurrence network. Radial position encodes character prominence; angular position reflects community structure. The embedding recovers the novel's multi-plot organization.}
    \label{fig:poincare}
\end{figure}

We extend geometric intuitions of character networks by learning Poincar\'{e} disk embeddings \citep{nickel2017poincare}, where hyperbolic space naturally encodes hierarchy: central characters embed near the origin, peripheral characters near the boundary. This allows us to conveniently visualize the narrative and community structure, see Figure \ref{fig:poincare} for visualization. We utilize the \texttt{gensim} software package \cite{rehurek_lrec} to generate Poincar\'{e} embeddings, and we set the embedding dimension to \(d=2\) and utilized negative sampling with \(k=10\) negative samples for each positive relation to optimize the loss function. Training was performed for \(100\) epochs for each embedding.

\begin{table*}
\footnotesize
\begin{longtable}{llc}
\caption{Corpus of 64 nineteenth-century British novels (1801--1900).}\label{tab:corpus}\\
\toprule
\textbf{Author} & \textbf{Title} & \textbf{Year} \\
\midrule
\endfirsthead
\multicolumn{3}{c}{\tablename\ \thetable{} -- continued from previous page} \\
\toprule
\textbf{Author} & \textbf{Title} & \textbf{Year} \\
\midrule
\endhead
\midrule
\multicolumn{3}{r}{Continued on next page} \\
\endfoot
\bottomrule
\endlastfoot
Maria Edgeworth & \textit{Belinda} & 1801 \\
Jane Austen & \textit{Sense and Sensibility} & 1811 \\
Jane Austen & \textit{Pride and Prejudice} & 1813 \\
Jane Austen & \textit{Mansfield Park} & 1814 \\
Walter Scott & \textit{Waverley} & 1814 \\
Jane Austen & \textit{Emma} & 1815 \\
Walter Scott & \textit{Guy Mannering} & 1815 \\
Walter Scott & \textit{Old Mortality} & 1816 \\
Jane Austen & \textit{Persuasion} & 1817 \\
Walter Scott & \textit{The Heart of Midlothian} & 1818 \\
Charles Maturin & \textit{Melmoth the Wanderer} & 1820 \\
Walter Scott & \textit{The Pirate} & 1821 \\
Benjamin Disraeli & \textit{Vivian Grey} & 1826 \\
Edward Bulwer-Lytton & \textit{Paul Clifford} & 1830 \\
William Harrison Ainsworth & \textit{Rookwood} & 1834 \\
Frederick Marryat & \textit{Mr.\ Midshipman Easy} & 1836 \\
Charles Dickens & \textit{Oliver Twist} & 1837 \\
William Harrison Ainsworth & \textit{Jack Sheppard} & 1839 \\
Harriet Martineau & \textit{Deerbrook} & 1839 \\
Frances Trollope & \textit{The Life and Adventures of Michael Armstrong, the Factory Boy} & 1840 \\
Benjamin Disraeli & \textit{Sybil} & 1845 \\
Charles Dickens & \textit{Dombey and Son} & 1846 \\
William Makepeace Thackeray & \textit{Vanity Fair} & 1847 \\
George W.\ M.\ Reynolds & \textit{The Mysteries of London} & 1848 \\
William Makepeace Thackeray & \textit{Pendennis} & 1848 \\
Elizabeth Gaskell & \textit{Mary Barton} & 1848 \\
Charlotte Bront\"{e} & \textit{Shirley} & 1849 \\
Elizabeth Gaskell & \textit{Ruth} & 1853 \\
Charlotte Yonge & \textit{The Heir of Redclyffe} & 1853 \\
Charles Dickens & \textit{Hard Times} & 1854 \\
Elizabeth Gaskell & \textit{North and South} & 1854 \\
Charles Kingsley & \textit{Westward Ho!} & 1855 \\
Anthony Trollope & \textit{Barchester Towers} & 1857 \\
Charles Dickens & \textit{A Tale of Two Cities} & 1859 \\
George Eliot & \textit{Adam Bede} & 1859 \\
George Eliot & \textit{The Mill on the Floss} & 1860 \\
Ellen Wood & \textit{East Lynne} & 1861 \\
Mary Elizabeth Braddon & \textit{Lady Audley's Secret} & 1862 \\
Mary Elizabeth Braddon & \textit{Aurora Floyd} & 1863 \\
George Eliot & \textit{Romola} & 1863 \\
Charles Reade & \textit{Hard Cash} & 1863 \\
Anthony Trollope & \textit{The Small House at Allington} & 1864 \\
Anthony Trollope & \textit{Can You Forgive Her?} & 1864 \\
Charles Dickens & \textit{Our Mutual Friend} & 1864 \\
Wilkie Collins & \textit{Armadale} & 1864 \\
Margaret Oliphant & \textit{Miss Marjoribanks} & 1866 \\
George Eliot & \textit{Middlemarch} & 1871 \\
Ouida & \textit{B\'{e}b\'{e}e} & 1874 \\
Margaret Oliphant & \textit{Phoebe Junior} & 1876 \\
George Eliot & \textit{Daniel Deronda} & 1876 \\
Thomas Hardy & \textit{The Return of the Native} & 1878 \\
Thomas Hardy & \textit{The Mayor of Casterbridge} & 1886 \\
Oscar Wilde & \textit{The Picture of Dorian Gray} & 1890 \\
George Gissing & \textit{New Grub Street} & 1891 \\
Thomas Hardy & \textit{Tess of the d'Urbervilles} & 1891 \\
Rudyard Kipling & \textit{The Light that Failed} & 1891 \\
George Gissing & \textit{The Odd Women} & 1893 \\
George Moore & \textit{Esther Waters} & 1894 \\
George Du Maurier & \textit{Trilby} & 1894 \\
Ella Hepworth Dixon & \textit{The Story of a Modern Woman} & 1894 \\
Joseph Conrad & \textit{Almayer's Folly} & 1895 \\
Thomas Hardy & \textit{Jude the Obscure} & 1895 \\
H.\,G.\ Wells & \textit{The Invisible Man} & 1897 \\
H.\,G.\ Wells & \textit{Love and Mr.\ Lewisham} & 1900 \\
\bottomrule
\end{longtable}
\end{table*}

\begin{table*}[t]
\centering \small
    \begin{tabular}{cccccccc|c|cccccc|c}
    \toprule
    \multicolumn{2}{c}{\textbf{Method}} & \multicolumn{7}{c|}{Mean Absolute Error ($\downarrow$)} & \multicolumn{7}{c}{Correlation ($\uparrow$)}\\\midrule
         \tiny\faBook & \tiny\faTag & \textbf{N} & \textbf{A} & \textbf{C} & \textbf{I} & \textbf{DC} & \textbf{DN} & \textbf{avg.} & \textbf{N} & \textbf{A} & \textbf{C} & \textbf{I} & \textbf{DC} & \textbf{DN} & \textbf{avg.}\\\midrule\midrule

 &  & 3.92 & 10.31 & 5.92 & 8.21 & 6.60 & 1.76 & 6.12 & 0.57 & 0.65 & 0.66 & 0.67 & 0.46 & 0.53 & 0.42 \\
 & \tiny\checkmark & 3.53 & 8.24 & 5.35 & 6.94 & 6.68 & 1.87 & 5.43 & 0.61 & 0.68 & 0.55 & 0.76 & 0.27 & 0.52 & 0.60 \\
\tiny\checkmark &  & 7.74 & 8.40 & 4.30 & 6.80 & 5.89 & 2.70 & 5.97 & 0.68 & 0.66 & 0.66 & 0.67 & 0.48 & 0.61 & 0.45 \\
\tiny\checkmark & \tiny\checkmark & 5.05 & 7.75 & 4.10 & 7.11 & 5.92 & 4.04 & 5.66 & 0.78 & 0.74 & 0.70 & 0.80 & 0.46 & 0.58 & 0.71 \\
\midrule
\multicolumn{2}{c}{BookNLP} & 1.57 & 7.40 & 4.55 & 7.12 & 4.25 & 1.23 & 4.35 & 0.86 & 0.52 & 0.49 & 0.29 & 0.62 & 0.42 & 0.46 \\\midrule
\multicolumn{2}{c}{BookNLP \faHandPaper} & 1.57 & 4.89 & 3.17 & 4.16 & 3.48 & 1.18 & 3.08 & 0.86 & 0.89 & 0.82 & 0.91 & 0.80 & 0.56 & 0.86 \\
\multicolumn{2}{c}{SpanLLM} & 3.06 & 8.50 & 3.07 & 7.71 & 4.08 & 1.30 & 4.62 & 0.62 & 0.79 & 0.81 & 0.76 & 0.77 & 0.52 & 0.55 \\
\bottomrule
    \end{tabular}
    \caption{(\textbf{\textit{Jane Eyre}}) Mean Absolute Error (MAE) and Correlation scores for the predicted component counts averaged across characters and chapters for each method we compare. The initial four methods apply GPT-4o-mini at different levels of granularity for chapter-level counts. A check in the book column indicates the input chapter text is chunked. A check in the tag column indicates the model is asked about each component individually. The bottom two approaches are our span-level methods. The BookNLP method with the end indicates we manually intervened to correct the co-reference issue with one of Jane's mention clusters. The BookNLP score without this intervention is also shown.}
    \label{tab:accuracy_je}
\end{table*}

\begin{figure*}[t]
    \centering
    \includegraphics[width=\linewidth]{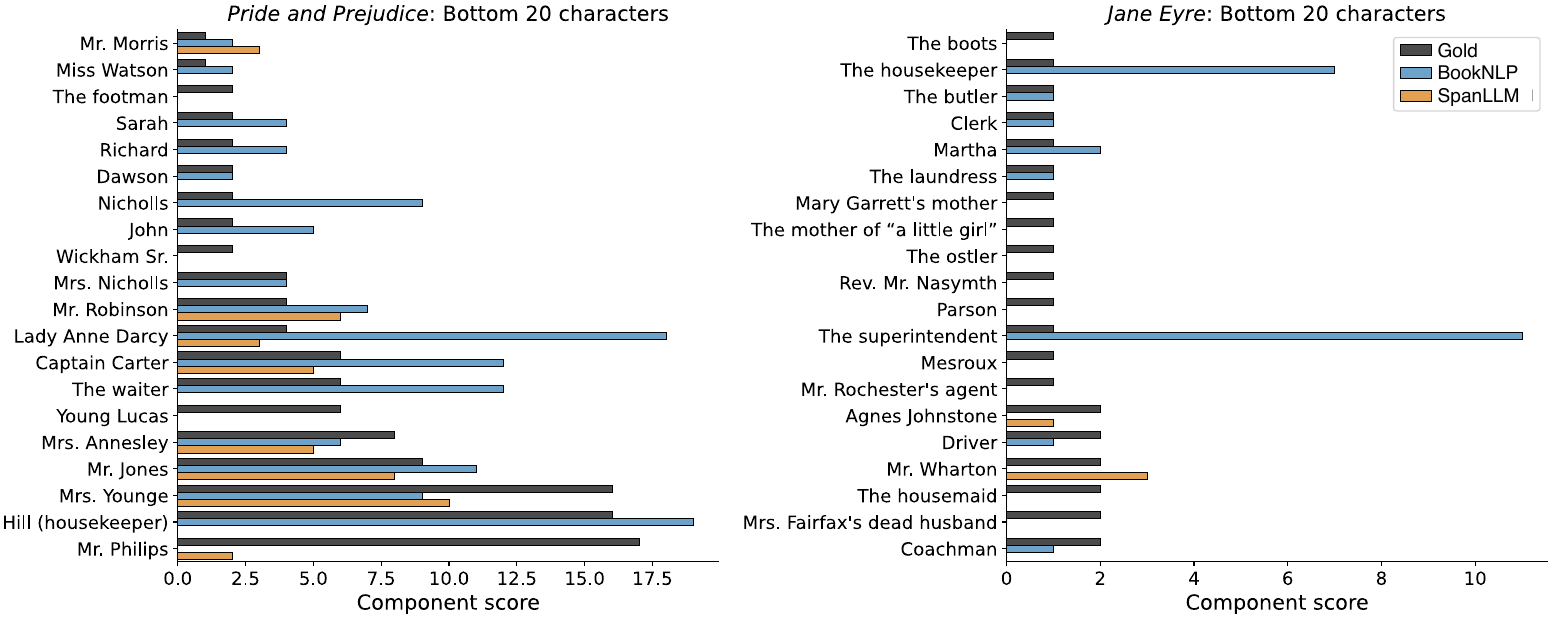}
    \caption{Plots showing the total component scores for our span-level methods on the bottom 20 characters in \textit{Pride and Prejudice} and \textit{Jane Eyre}. We see much greater variability in this set of characters than for the top characters, in part because the total counts are much lower. For \textit{Jane Eyre}, the BookNLP approach uses the manual correction of the co-reference issue for Jane.}
    \label{fig:bottomcounts}
\end{figure*}

\begin{figure*}[t]
    \centering
    \includegraphics[width=\linewidth]{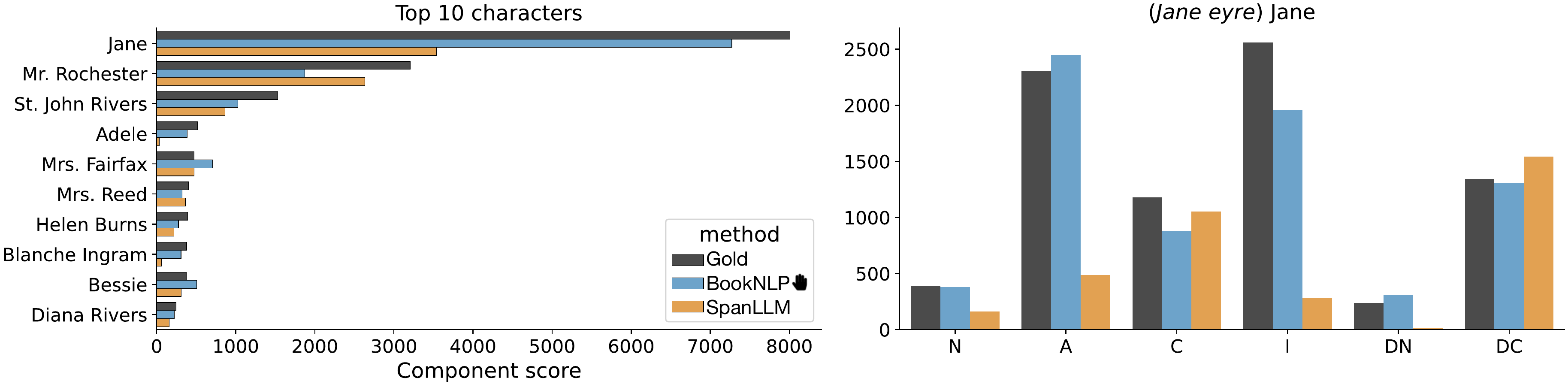}
    \caption{Plots showing component scores from our span-level methods for top characters in \textit{Jane Eyre}. The left plot shows total component score for all tags across the top 10 characters and the right plot shows a breakdown of tags by component for the protagonist, Jane.}
    \label{fig:topchars_je}
\end{figure*}

\begin{figure*}[t]
    \centering
    \includegraphics[width=\linewidth]{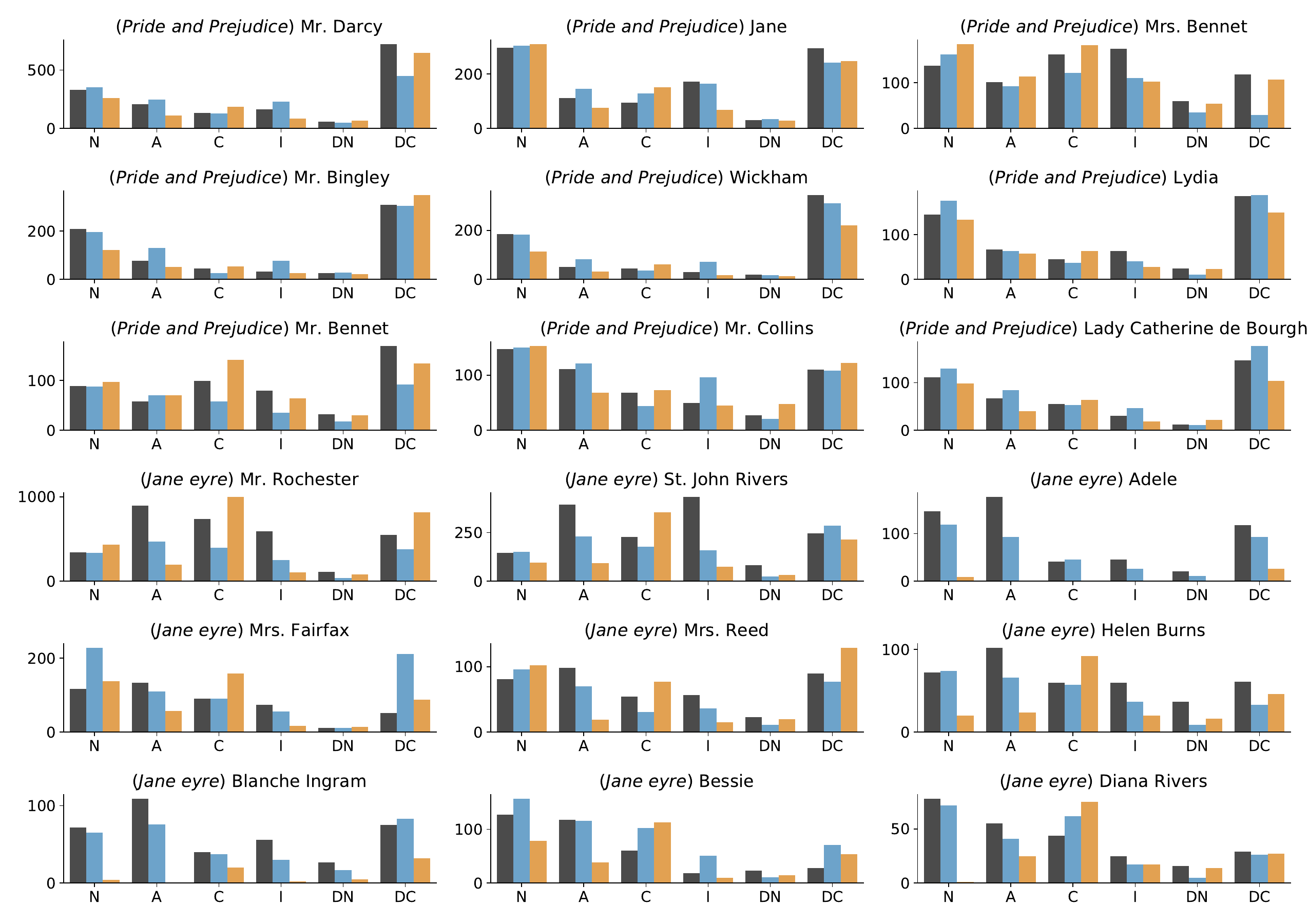}
    \caption{Plots showing total component scores broken down by tag for more of the top characters in each novel. The grey bar is the gold score, the blue bar is our BookNLP method (with manual correction of the co-reference issue for \textit{Jane Eyre}), and the orange bar is our SpanLLM method.}
    \label{fig:allchars}
\end{figure*}

\begin{figure*}[t]
\centering
\begin{subfigure}[t]{0.49\textwidth}
  \centering
  \includegraphics[width=\linewidth]{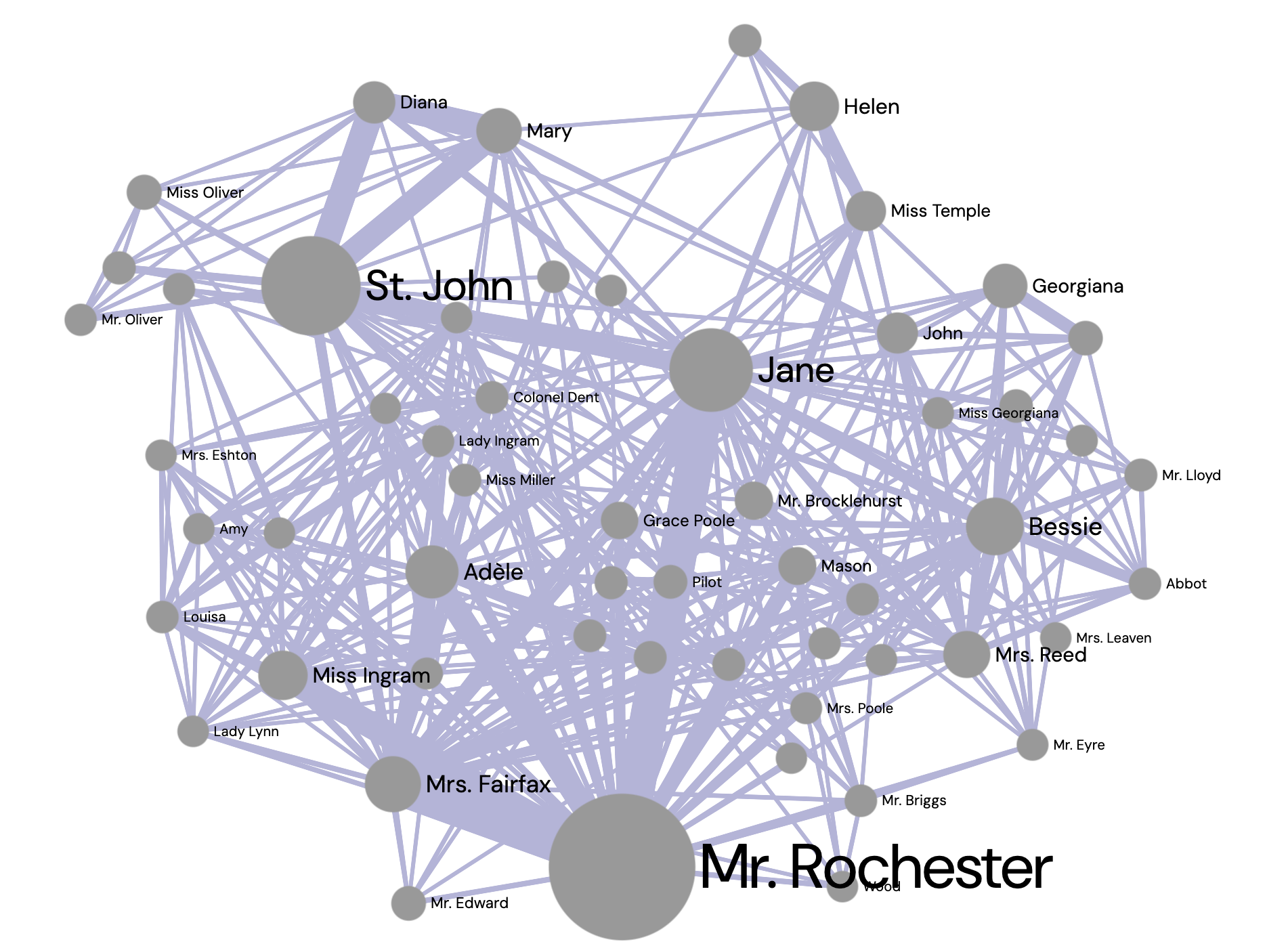}
  \caption{Raw co-occurrence output (narrator cluster not merged).}
  \label{fig:jane}
\end{subfigure}
\hfill
\begin{subfigure}[t]{0.49\textwidth}
  \centering
  \includegraphics[width=\linewidth]{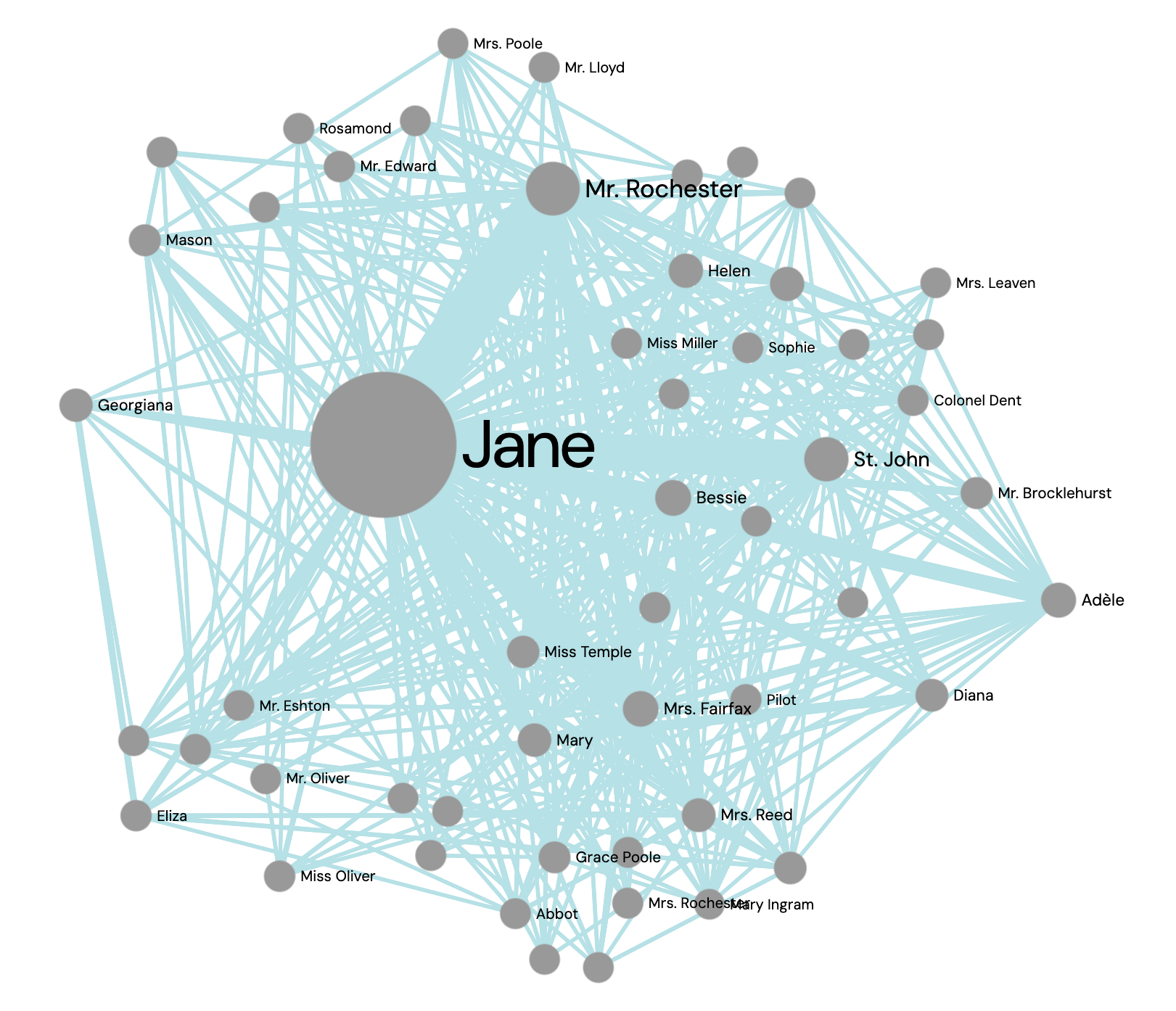}
  \caption{Narrator cluster merged into \texttt{Jane}.}
  \label{fig:jane-merged}
\end{subfigure}
\caption{Co-occurrence networks for \textit{Jane Eyre}. Nodes are sized by total mentions; edges denote paragraph-level co-occurrence (weighted). Merging the first-person narrator cluster restores Jane’s centrality and strengthens her ties to major characters.}
\label{fig:jane-ablation}
\end{figure*}

\begin{figure*}[h!]
\centering
\begin{subfigure}[t]{0.45\textwidth}
  \centering
  \includegraphics[width=\linewidth]{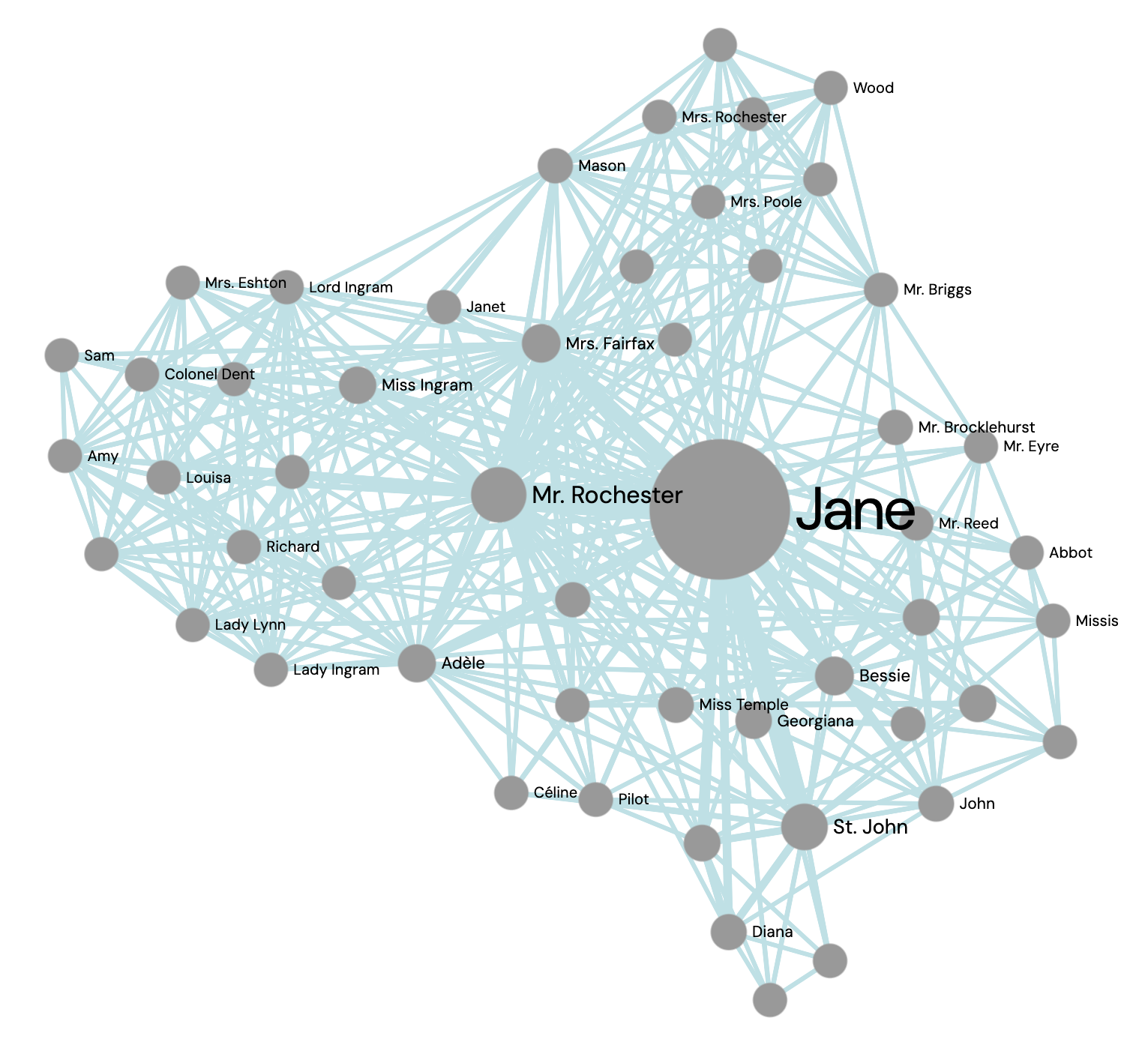}
  \caption{Co-occurrence Network.}
  \label{fig:jane}
\end{subfigure}
\hfill
\begin{subfigure}[t]{0.45\textwidth}
  \centering
  \includegraphics[width=\linewidth]{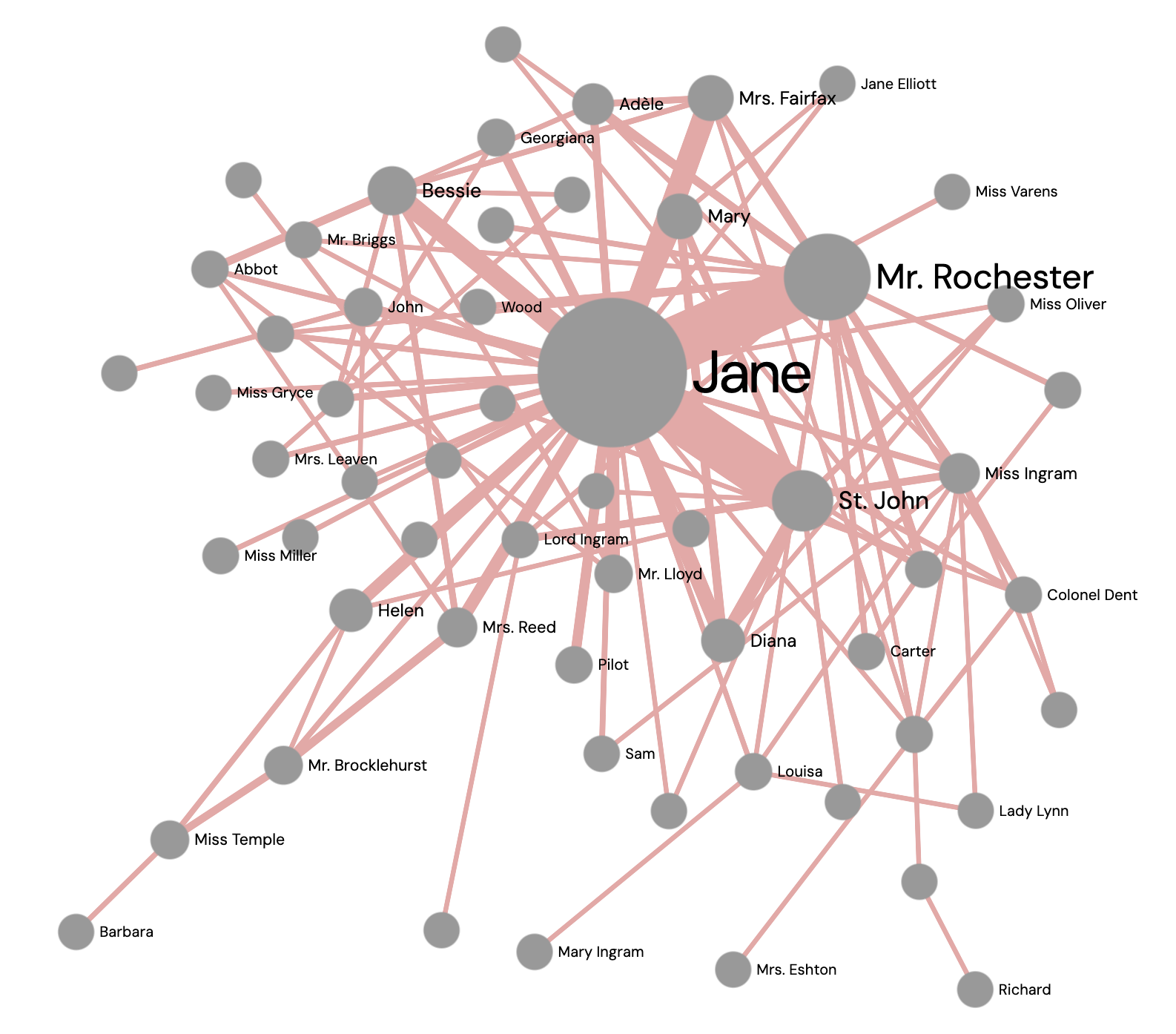}
  \caption{Dialogue Network (C).}
  \label{fig:jane_dia}
\end{subfigure}
\begin{subfigure}[t]{0.45\textwidth}
  \centering
  \includegraphics[width=\linewidth]{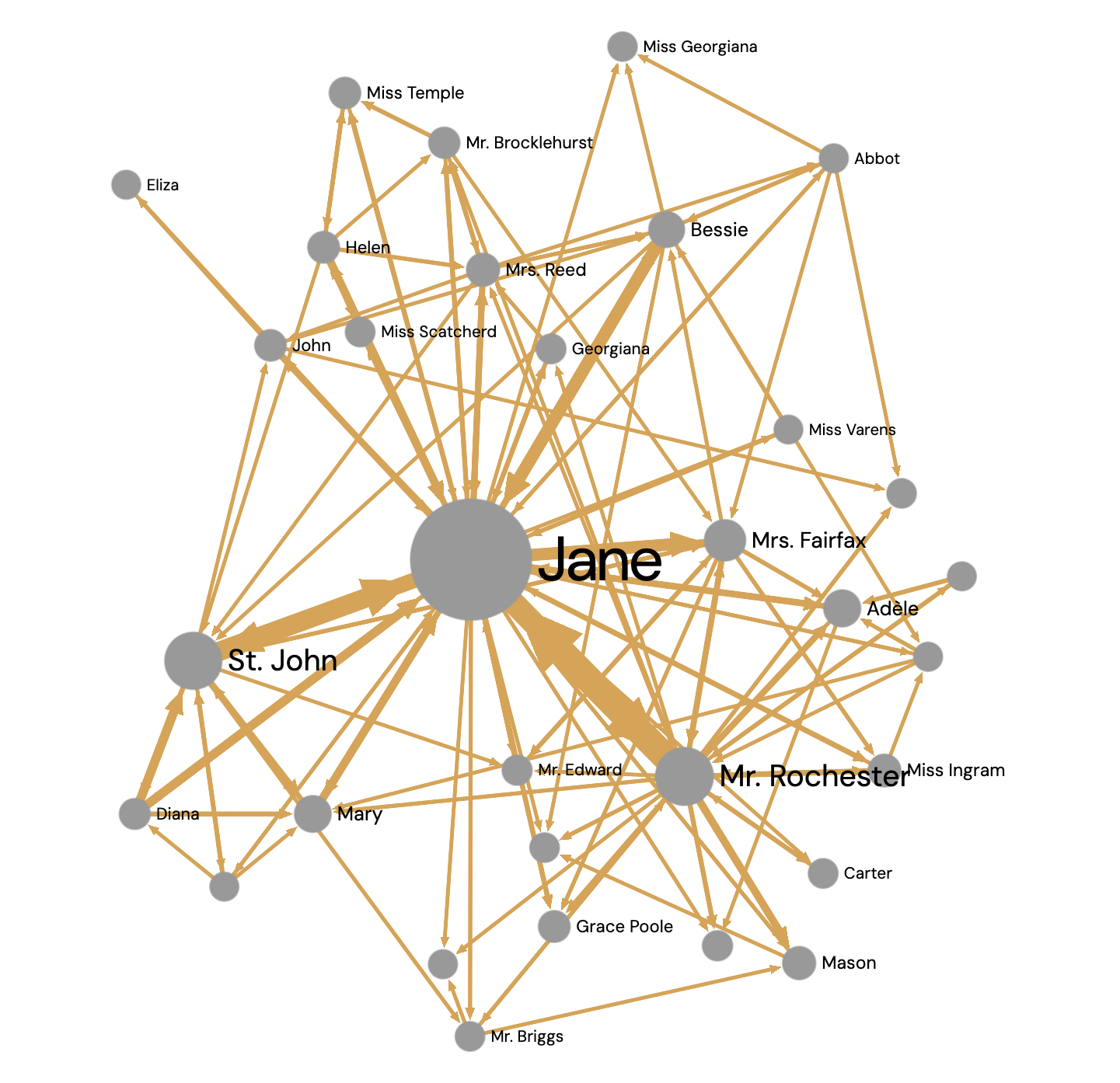}
  \caption{Discussion Network (DC).}
  \label{fig:jane_gossip}
\end{subfigure}
\caption{Character networks for \textit{Jane Eyre} illustrating the three edge representations chosen.}
\label{fig:more-jane}
\end{figure*}

\begin{table}[h]
\centering
\small
\caption{Corpus-averaged global network statistics (mean $\pm$ SD across 73 novels).}
\label{tab:corpus-summary}
\begin{tabular}{lccc}
\toprule
\textbf{Metric} & \textbf{Co-occurrence} & \textbf{Dialogue} & \textbf{Discussion (DC)} \\
\midrule
Nodes & 391 $\pm$ 279 & 66 $\pm$ 41 & 190 $\pm$ 121 \\
Edges & 1699 $\pm$ 1857 & 157 $\pm$ 101 & 393 $\pm$ 244 \\
Density & 0.04 $\pm$ 0.12 & 0.10 $\pm$ 0.06 & 0.02 $\pm$ 0.01 \\
Degree Gini & 0.61 $\pm$ 0.08 & 0.47 $\pm$ 0.07 & --- \\
Centralization & 0.43 $\pm$ 0.15 & 0.46 $\pm$ 0.17 & --- \\
Clustering & 0.64 $\pm$ 0.10 & 0.47 $\pm$ 0.12 & --- \\
Assortativity & $-$0.29 $\pm$ 0.09 & $-$0.31 $\pm$ 0.11 & --- \\
Avg Path Length & 2.53 $\pm$ 0.34 & 2.64 $\pm$ 0.78 & --- \\
\midrule
In-strength Gini & --- & --- & 0.76 $\pm$ 0.10 \\
Out-strength Gini & --- & --- & 0.72 $\pm$ 0.09 \\
Reciprocity & --- & --- & 0.25 $\pm$ 0.08 \\
\bottomrule
\end{tabular}
\end{table}

\onecolumn
\begin{longtable}{l|rrr|rrr|rrr}
\caption{Network statistics for all novels in the corpus.}
\label{tab:corpus-full}\\
\toprule
& \multicolumn{3}{c|}{\textbf{Co-occurrence}} & \multicolumn{3}{c|}{\textbf{Dialogue}} & \multicolumn{3}{c}{\textbf{Discussion (DC)}} \\
\textbf{Novel} & $|V|$ & Gini & Cent. & $|V|$ & Gini & Cent. & $|V|$ & Gini$_{in}$ & Recip. \\
\midrule
\endfirsthead
\multicolumn{10}{c}{\tablename\ \thetable{} -- continued from previous page} \\
\toprule
& \multicolumn{3}{c|}{\textbf{Co-occurrence}} & \multicolumn{3}{c|}{\textbf{Dialogue}} & \multicolumn{3}{c}{\textbf{Discussion (DC)}} \\
\textbf{Novel} & $|V|$ & Gini & Cent. & $|V|$ & Gini & Cent. & $|V|$ & Gini$_{in}$ & Recip. \\
\midrule
\endhead
\midrule
\multicolumn{10}{r}{Continued on next page} \\
\endfoot
\bottomrule
\endlastfoot
\textit{A Tale of Two Cities} & 175 & 0.66 & 0.28 & 34 & 0.48 & 0.39 & 79 & 0.72 & 0.34 \\
\textit{Adam Bede} & 309 & 0.62 & 0.46 & 63 & 0.51 & 0.38 & 208 & 0.81 & 0.20 \\
\textit{Almayer's Folly} & 92 & 0.57 & 0.63 & 23 & 0.41 & 0.39 & 47 & 0.72 & 0.30 \\
\textit{Armadale} & 279 & 0.65 & 0.46 & 34 & 0.46 & 0.56 & 168 & 0.85 & 0.20 \\
\textit{Aurora Floyd (Vol.\ 1)} & 216 & 0.56 & 0.59 & 22 & 0.40 & 0.56 & 58 & 0.70 & 0.29 \\
\textit{Aurora Floyd (Vol.\ 2)} & 184 & 0.57 & 0.40 & 29 & 0.43 & 0.44 & 67 & 0.72 & 0.32 \\
\textit{Aurora Floyd (Vol.\ 3)} & 102 & 0.57 & 0.52 & 24 & 0.44 & 0.51 & 48 & 0.76 & 0.35 \\
\textit{Barchester Towers} & 373 & 0.61 & 0.34 & 64 & 0.50 & 0.38 & 123 & 0.77 & 0.28 \\
\textit{Belinda} & 365 & 0.64 & 0.59 & 47 & 0.50 & 0.45 & 276 & 0.85 & 0.26 \\
\textit{B\'{e}b\'{e}e} & 130 & 0.59 & 0.58 & 17 & 0.38 & 0.71 & 47 & 0.66 & 0.34 \\
\textit{Can You Forgive Her?} & 470 & 0.65 & 0.44 & 78 & 0.52 & 0.29 & 235 & 0.85 & 0.31 \\
\textit{Daniel Deronda} & 588 & 0.63 & 0.41 & 107 & 0.52 & 0.40 & 311 & 0.83 & 0.23 \\
\textit{Deerbrook} & 235 & 0.66 & 0.51 & 47 & 0.47 & 0.51 & 144 & 0.81 & 0.35 \\
\textit{Dombey and Son} & 466 & 0.67 & 0.41 & 92 & 0.54 & 0.39 & 241 & 0.81 & 0.26 \\
\textit{East Lynne} & 303 & 0.66 & 0.51 & 77 & 0.53 & 0.56 & 202 & 0.84 & 0.33 \\
\textit{Emma} & 322 & 0.64 & 0.60 & 45 & 0.53 & 0.64 & 186 & 0.84 & 0.25 \\
\textit{Esther Waters} & 240 & 0.60 & 0.57 & 58 & 0.52 & 0.65 & 143 & 0.80 & 0.27 \\
\textit{Guy Mannering} & 649 & 0.66 & 0.18 & 99 & 0.52 & 0.27 & 250 & 0.70 & 0.18 \\
\textit{Hard Cash} & 746 & 0.65 & 0.37 & 114 & 0.54 & 0.51 & 372 & 0.78 & 0.22 \\
\textit{Hard Times} & 179 & 0.63 & 0.41 & 21 & 0.37 & 0.56 & 92 & 0.81 & 0.32 \\
\textit{Jack Sheppard} & 373 & 0.62 & 0.42 & 77 & 0.54 & 0.48 & 228 & 0.80 & 0.30 \\
\textit{Jude the Obscure} & 247 & 0.58 & 0.52 & 23 & 0.46 & 0.56 & 110 & 0.86 & 0.17 \\
\textit{Lady Audley's Secret} & 262 & 0.61 & 0.57 & 42 & 0.50 & 0.75 & 111 & 0.78 & 0.25 \\
\textit{Love and Mr. Lewisham} & 168 & 0.59 & 0.58 & 22 & 0.45 & 0.86 & 86 & 0.71 & 0.27 \\
\textit{Mansfield Park} & 200 & 0.61 & 0.66 & 28 & 0.41 & 0.45 & 139 & 0.83 & 0.35 \\
\textit{Mary Barton} & 298 & 0.67 & 0.43 & 52 & 0.52 & 0.55 & 152 & 0.80 & 0.31 \\
\textit{Melmoth the Wanderer (1)} & 179 & 0.50 & 0.27 & 29 & 0.33 & 0.28 & 33 & 0.40 & 0.13 \\
\textit{Melmoth the Wanderer (2)} & 84 & 0.53 & 0.40 & 17 & 0.24 & 0.25 & 13 & 0.26 & 0.00 \\
\textit{Melmoth the Wanderer (3)} & 157 & 0.64 & 0.29 & 37 & 0.25 & 0.16 & 67 & 0.66 & 0.09 \\
\textit{Melmoth the Wanderer (4)} & 197 & 0.52 & 0.26 & 38 & 0.34 & 0.16 & 71 & 0.68 & 0.15 \\
\textit{Middlemarch} & 678 & 0.64 & 0.35 & 136 & 0.54 & 0.30 & 415 & 0.83 & 0.24 \\
\textit{Miss Marjoribanks} & 198 & 0.58 & 0.81 & 52 & 0.52 & 0.69 & 117 & 0.81 & 0.34 \\
\textit{Mr. Midshipman Easy} & 242 & 0.64 & 0.58 & 54 & 0.49 & 0.77 & 159 & 0.76 & 0.33 \\
\textit{New Grub Street} & 232 & 0.65 & 0.37 & 31 & 0.45 & 0.49 & 140 & 0.81 & 0.29 \\
\textit{North and South} & 291 & 0.67 & 0.60 & 71 & 0.55 & 0.73 & 187 & 0.83 & 0.22 \\
\textit{Old Mortality} & 738 & 0.60 & 0.32 & 130 & 0.49 & 0.27 & 314 & 0.73 & 0.17 \\
\textit{Oliver Twist}  & 180 & 0.67 & 0.46 & 42 & 0.40 & 0.48 & 94 & 0.73 & 0.38 \\
\textit{Our Mutual Friend} & 558 & 0.64 & 0.32 & 103 & 0.51 & 0.29 & 350 & 0.82 & 0.26 \\
\textit{Paul Clifford} & 576 & 0.64 & 0.29 & 80 & 0.50 & 0.30 & 228 & 0.72 & 0.17 \\
\textit{Pendennis} & 1501 & 0.59 & 0.50 & 169 & 0.55 & 0.46 & 615 & 0.75 & 0.18 \\
\textit{Persuasion} & 131 & 0.57 & 0.70 & 27 & 0.41 & 0.51 & 85 & 0.76 & 0.37 \\
\textit{Phoebe Junior} & 220 & 0.62 & 0.55 & 35 & 0.49 & 0.58 & 141 & 0.82 & 0.23 \\
\textit{Pride and Prejudice} & 187 & 0.58 & 0.58 & 37 & 0.45 & 0.54 & 107 & 0.82 & 0.31 \\
\textit{Romola} & 507 & 0.63 & 0.40 & 54 & 0.48 & 0.36 & 306 & 0.76 & 0.15 \\
\textit{Rookwood} & 572 & 0.65 & 0.31 & 90 & 0.54 & 0.32 & 228 & 0.81 & 0.26 \\
\textit{Ruth} & 227 & 0.62 & 0.61 & 33 & 0.45 & 0.72 & 128 & 0.81 & 0.30 \\
\textit{Sense and Sensibility} & 150 & 0.63 & 0.52 & 27 & 0.44 & 0.62 & 101 & 0.81 & 0.31 \\
\textit{Shirley}  & 498 & 0.66 & 0.30 & 78 & 0.53 & 0.33 & 279 & 0.81 & 0.21 \\
\textit{Sybil}  & 532 & 0.62 & 0.20 & 86 & 0.47 & 0.28 & 286 & 0.71 & 0.20 \\
\textit{Tess of the d'Urbervilles} & 274 & 0.63 & 0.49 & 38 & 0.51 & 0.75 & 79 & 0.78 & 0.28 \\
\textit{The Heart of Midlothian (1)} & 438 & 0.63 & 0.22 & 78 & 0.46 & 0.23 & 174 & 0.66 & 0.16 \\
\textit{The Heart of Midlothian (2)} & 565 & 0.60 & 0.35 & 81 & 0.45 & 0.54 & 267 & 0.64 & 0.14 \\
\textit{The Heir of Redclyffe} & 384 & 0.65 & 0.58 & 83 & 0.56 & 0.46 & 274 & 0.88 & 0.26 \\
\textit{The Invisible Man} & 94 & 0.62 & 0.22 & 26 & 0.38 & 0.40 & 41 & 0.60 & 0.30 \\
\textit{Michael Armstrong} & 288 & 0.66 & 0.38 & 77 & 0.51 & 0.37 & 141 & 0.81 & 0.32 \\
\textit{The Light that Failed} & 167 & 0.61 & 0.59 & 32 & 0.46 & 0.80 & 100 & 0.83 & 0.20 \\
\textit{The Mayor of Casterbridge} & 185 & 0.63 & 0.56 & 33 & 0.48 & 0.62 & 75 & 0.76 & 0.27 \\
\textit{The Mill on the Floss} & 363 & 0.62 & 0.57 & 57 & 0.55 & 0.55 & 222 & 0.85 & 0.25 \\
\textit{The Mysteries of London (1)}  & 588 & 0.67 & 0.20 & 107 & 0.50 & 0.34 & 267 & 0.78 & 0.27 \\
\textit{The Mysteries of London (2)}  & 614 & 0.63 & 0.21 & 149 & 0.51 & 0.32 & 300 & 0.79 & 0.28 \\
\textit{The Mysteries of London (3)}  & 579 & 0.67 & 0.19 & 110 & 0.51 & 0.21 & 321 & 0.79 & 0.30 \\
\textit{The Mysteries of London (4)}  & 606 & 0.63 & 0.20 & 130 & 0.45 & 0.20 & 295 & 0.73 & 0.23 \\
\textit{The Odd Women}  & 155 & 0.65 & 0.34 & 35 & 0.48 & 0.58 & 96 & 0.81 & 0.37 \\
\textit{The Picture of Dorian Gray} & 245 & 0.51 & 0.67 & 28 & 0.39 & 0.37 & 111 & 0.83 & 0.16 \\
\textit{The Pirate} & 863 & 0.73 & 0.20 & 170 & 0.53 & 0.22 & 335 & 0.73 & 0.20 \\
\textit{The Return of the Native} & 214 & 0.68 & 0.35 & 29 & 0.44 & 0.42 & 85 & 0.81 & 0.35 \\
\textit{The Small House at Allington}  & 387 & 0.65 & 0.37 & 74 & 0.51 & 0.38 & 211 & 0.82 & 0.31 \\
\textit{The Story of a Modern Woman}  & 154 & 0.00 & 0.00 & 36 & 0.43 & 0.46 & 58 & 0.73 & 0.25 \\
\textit{Trilby} & 514 & 0.60 & 0.32 & 81 & 0.47 & 0.45 & 199 & 0.70 & 0.12 \\
\textit{Vanity Fair}  & 1363 & 0.58 & 0.27 & 175 & 0.52 & 0.23 & 450 & 0.74 & 0.20 \\
\textit{Vivian Grey}  & 737 & 0.60 & 0.58 & 98 & 0.43 & 0.64 & 400 & 0.69 & 0.13 \\
\textit{Waverley} & 1023 & 0.64 & 0.43 & 141 & 0.42 & 0.39 & 327 & 0.65 & 0.12 \\
\textit{Westward Ho!} & 945 & 0.64 & 0.44 & 123 & 0.56 & 0.53 & 484 & 0.72 & 0.14 \\
\end{longtable}

\clearpage
\section{Prompts}
Prompts are expressed as f-strings or a brief code snippet including an f-string. The values inserted into the strings should be intuitive but you can reference the codebase for any clarifications. 
\label{sec:prompts}
\begin{tcolorbox}[
    colback=white,    
    colframe=brown,  
    coltitle=white,   
    fonttitle=\bfseries, 
    title=Prompt for counting all tags in a chunk of text,
    fontupper=\ttfamily
]
\small
\begin{Verbatim}[breaklines=true,breaksymbolleft={},breaksymbolright={},tabsize=0]
    Please analyze the following text and tag all instances of the following characters:
    {', '.join(character_set)}

    For each character, include all name variants under the normalized form above (e.g., tags for "Lizzy" should be counted under "Elizabeth")

    For each character, provide the following tags:
    - N: Count all named mentions of this person (including name variants).
    - A: Count each verb of physical action (exclude speech, thought, and feeling verbs).
    - C: Count blocks of directly quoted dialogue, paraphrased speech, and letters.
    - I: Count each verb expressing thought, feeling, intention, or interpretation.
    - DN: Count sentences describing the character by the narrator.
    - DC: Count sentences where other characters discuss the character.
    
    Text:
    {text}

    Return the results as a JSON object with each character as a key and their tags as values. Only include characters that have at least one non-zero tag.
    If a character has all tags equal to zero, omit it from the JSON output entirely.

    Strictly output JSON in this format:
    {{
        "Character1": {{"N": count1, "A": countA1, "C": countC1, "I": countI1, "DN": countDN1, "DC": countDC1}},
        "Character2": {{"N": count2, "A": countA2, "C": countC2, "I": countI2, "DN": countDN2, "DC": countDC2}},
        ...
    }}
\end{Verbatim}
\end{tcolorbox}

\begin{figure*}
\begin{tcolorbox}[
    colback=white,    
    colframe=brown,  
    coltitle=white,   
    fonttitle=\bfseries, 
    title=Prompt for counting an individual tag in a chunk of text,
    fontupper=\ttfamily
]
\small
\begin{Verbatim}[breaklines=true,breaksymbolleft={},breaksymbolright={},tabsize=0]
    descriptions = {"N": "Count all named mentions of this person (including name variants).",
                    "A": "Count each verb of physical action (exclude speech, thought, and feeling verbs).",
                    "C": "Count blocks of directly quoted dialogue, paraphrased speech, and letters.",
                    "I": "Count each verb expressing thought, feeling, intention, or interpretation.",
                    "DN": "Count sentences describing the character by the narrator.",
                    "DC": "Count sentences where other characters discuss the character."
                    }

    prompt = f"""
    Please analyze the following text and tag all instances of the following characters:
    {', '.join(character_set)}

    For each character, include all name variants under the normalized form above (e.g., tags for "Lizzy" should be counted under "Elizabeth")

    Provide the counts for the '{tag}' tag only.

    The tag '{tag}' is defined as:
    - {descriptions[tag]}
    
    Text:
    {text}

    Return the results as a JSON object with each character as a key and its '{tag}' count as value. Only include characters that have a non-zero count.
    If a character has '{tag}' count equal to zero, omit it from the JSON output entirely.

    Strictly output JSON in this format:
    {{
        "Character1": count1,
        "Character2": count2,
        ...
    }}
    """
\end{Verbatim}
\end{tcolorbox}
\end{figure*}

\begin{figure*}
\begin{tcolorbox}[
    colback=white,    
    colframe=brown,  
    coltitle=white,   
    fonttitle=\bfseries, 
    title=Prompt for limiting the character list to those present in a chapter,
    fontupper=\ttfamily
]
\small
\begin{Verbatim}[breaklines=true,breaksymbolleft={},breaksymbolright={},tabsize=0]
    {booktitle} section: {text}
    
    Character list: {', '.join(character_set)}
    
    Which characters from the character list are present in the {booktitle} section? Include any character that are mentioned or referred to, even if they are not physically present in the scenes.
    
    Strictly output a list of the characters who are present in this format: [character1, character2, etc.]
\end{Verbatim}
\end{tcolorbox}
\end{figure*}

\begin{figure*}
\begin{tcolorbox}[
    colback=white,    
    colframe=brown,  
    coltitle=white,   
    fonttitle=\bfseries, 
    title=Prompt for determining if a BookNLP character is correct,
    fontupper=\ttfamily
]
\small
\begin{Verbatim}[breaklines=true,breaksymbolleft={},breaksymbolright={},tabsize=0]
    Consider the full text of {book} and all the characters it contains. Each character can be referred to by multiple different variants of ther name. Is {character_name} a character in {book}? Answer yes or no.

    Answer:
\end{Verbatim}
\end{tcolorbox}
\end{figure*}

\begin{figure*}
\begin{tcolorbox}[
    colback=white,    
    colframe=brown,  
    coltitle=white,   
    fonttitle=\bfseries, 
    title=Prompt for mapping BookNLP character names to our gold character names,
    fontupper=\ttfamily
]
\small
\begin{Verbatim}[breaklines=true,breaksymbolleft={},breaksymbolright={},tabsize=0]
    Consider the following list of characters in {book}:
    {character_set}

    Which character from this list does the name {character_name} refer to? Answer with just the character name.

    Answer:
\end{Verbatim}
\end{tcolorbox}
\end{figure*}

\begin{figure*}
\begin{tcolorbox}[
    colback=white,    
    colframe=brown,  
    coltitle=white,   
    fonttitle=\bfseries, 
    title=Prompt for span-level DN,
    fontupper=\ttfamily
]
\small
\begin{Verbatim}[breaklines=true,breaksymbolleft={},breaksymbolright={},tabsize=0]
    Please analyze the following text for places where the narrator describes the following characters:
    {', '.join(character_set)}

    For each character, list all instances where the narration describes something about the character's looks, manner, or dress. If a character is not described, do not include them.

    Text:
    {text}

    Return the results as a JSON object with each character as a key and the list of descriptions as the value. Do not include instances of the character speaking, only descriptions from the narrator.

    Strictly output JSON in this format:
    {{
        "Character1": [phrase1, phrase2, ...],
        "Character2": [phrase1, ...],
        ...
    }}
\end{Verbatim}
\end{tcolorbox}
\end{figure*}

\begin{figure*}
\begin{tcolorbox}[
    colback=white,    
    colframe=brown,  
    coltitle=white,   
    fonttitle=\bfseries, 
    title=Prompt for span-level DC,
    fontupper=\ttfamily
]
\small
\begin{Verbatim}[breaklines=true,breaksymbolleft={},breaksymbolright={},tabsize=0]
    Please analyze the following dialogue for mentions of the following characters:
    {', '.join(character_set)}
        
    Surrounding text for context:
    {chunk}
    
    Dialogue sentence (speaker: {character_name}):
    {dialogue_sentence}
    
    Resolve any pronouns, relational mentions, or name variants in the dialogue sentence to the proper character name.
    
    Give your response as a JSON object with pronouns/mentions/names as the key and the character it refers to as the value in this format:
    {{
        "pronoun": character1,
        "pronoun": character2,
        ...
    }}
\end{Verbatim}
\end{tcolorbox}
\end{figure*}

\begin{figure*}
\begin{tcolorbox}[
    colback=white,    
    colframe=brown,  
    coltitle=white,   
    fonttitle=\bfseries, 
    title=Prompt for span-level C,
    fontupper=\ttfamily
]
\small
\begin{Verbatim}[breaklines=true,breaksymbolleft={},breaksymbolright={},tabsize=0]
    Please analyze the dialogue and letters in the following text.
    
    Full text for context:
    {text}
    
    Dialogue turn (or letter):{dialogue_turn}
    
    Character list: {character_set}
    
    Which character from the character list is the speaker (or writer) of this dialogue turn (or letter)?
    
    Give just the character name as your response.
    
    Answer:
\end{Verbatim}
\end{tcolorbox}
\end{figure*}

\begin{figure*}
\begin{tcolorbox}[
    colback=white,    
    colframe=brown,  
    coltitle=white,   
    fonttitle=\bfseries, 
    title=Prompt for span-level I,
    fontupper=\ttfamily
]
\small
\begin{Verbatim}[breaklines=true,breaksymbolleft={},breaksymbolright={},tabsize=0]
    Please analyze the following text for descriptions of the feelings and thoughts of the following characters:
    {', '.join(character_set)}

    For each character, list all instances where the narration shows us the characters thoughts, feelings, intentions, or perceptions.

    Text:
    {text}

    Return the results as a JSON object with each character as a key and the list of thoughts/feelings as the value. Do not include instances of the character speaking, only descriptions from the narrator.

    Strictly output JSON in this format:
    {{
        "Character1": [phrase1, phrase2, ...],
        "Character2": [phrase1, ...],
        ...
    }}
\end{Verbatim}
\end{tcolorbox}
\end{figure*}

\begin{figure*}
\begin{tcolorbox}[
    colback=white,    
    colframe=brown,  
    coltitle=white,   
    fonttitle=\bfseries, 
    title=Prompt for span-level A,
    fontupper=\ttfamily
]
\small
\begin{Verbatim}[breaklines=true,breaksymbolleft={},breaksymbolright={},tabsize=0]
    Please analyze the following text for actions performed by the following characters:
    {', '.join(character_set)}

    For each character, list all physical actions they perform. Do not include acts of talking, speaking, asking questions, or writing. Do not include acts of thinking or feeling.

    Text:
    {text}

    Return the results as a JSON object with each character as a key and the list of actions as the value.

    Strictly output JSON in this format:
    {{
        "Character1": [phrase1, phrase2, ...],
        "Character2": [phrase1, ...],
        ...
    }}
\end{Verbatim}
\end{tcolorbox}
\end{figure*}

\begin{figure*}
\begin{tcolorbox}[
    colback=white,    
    colframe=brown,  
    coltitle=white,   
    fonttitle=\bfseries, 
    title=Prompt for span-level N,
    fontupper=\ttfamily
]
\small
\begin{Verbatim}[breaklines=true,breaksymbolleft={},breaksymbolright={},tabsize=0]
    Please analyze the following text for mentions of the following characters:
    {', '.join(character_set)}

    For each character, list all variants of their name that are used to refer to them in the text. Do not include pronouns, only proper names. If they are referred to with and without a title, include both variants.

    Text:
    {text}

    Return the results as a JSON object with each character as a key and the list of name variants as the value.

    Strictly output JSON in this format:
    {{
        "Character1": [name1, name2, ...],
        "Character2": [name1, ...],
        ...
    }}
\end{Verbatim}
\end{tcolorbox}
\end{figure*}

\end{document}